\documentclass[10pt,twocolumn,letterpaper]{article}

\usepackage{cvpr}              % To produce the CAMERA-READY version
\definecolor{cvprblue}{rgb}{0.21,0.49,0.74}
\usepackage[pagebackref,breaklinks,colorlinks,allcolors=cvprblue]{hyperref}

\usepackage{wrapfig}

\def\confName{CVPR}
\def\confYear{2026}

\usepackage{multirow}
\usepackage{array}
\usepackage[capitalize]{cleveref}
\usepackage{algorithm}
\usepackage{algorithmic}
\usepackage[table]{xcolor}
\usepackage{etoc}

\addtocontents{toc}{\protect\setcounter{tocdepth}{-1}} 

\title{Self-Corrected Image Generation with Explainable Latent Rewards\vspace{-3mm}}
\newcommand{\method}{xLARD\xspace}

\author{%
  Yinyi Luo${}^{1,2}$\ , Hrishikesh Gokhale${}^1$, Marios Savvides${}^1$, Jindong Wang${}^3$,
  Shengfeng He${}^2$\footnotemark[2] \\
  ${}^1$Carnegie Mellon University, ${}^2$Singapore Management University, ${}^3$William \& Mary\\
  \small{\texttt{yinyil@andrew.cmu.edu}, \texttt{shengfenghe@smu.edu.sg}} \\
}
\begin{document}

\twocolumn[
\maketitle
\vspace{-17mm}\begin{center}
    \setlength{\fboxrule}{0pt}
    
    \fbox{%
  \begin{minipage}[t]{0.75\linewidth}
    \vspace{0pt}
    \includegraphics[width=\linewidth]{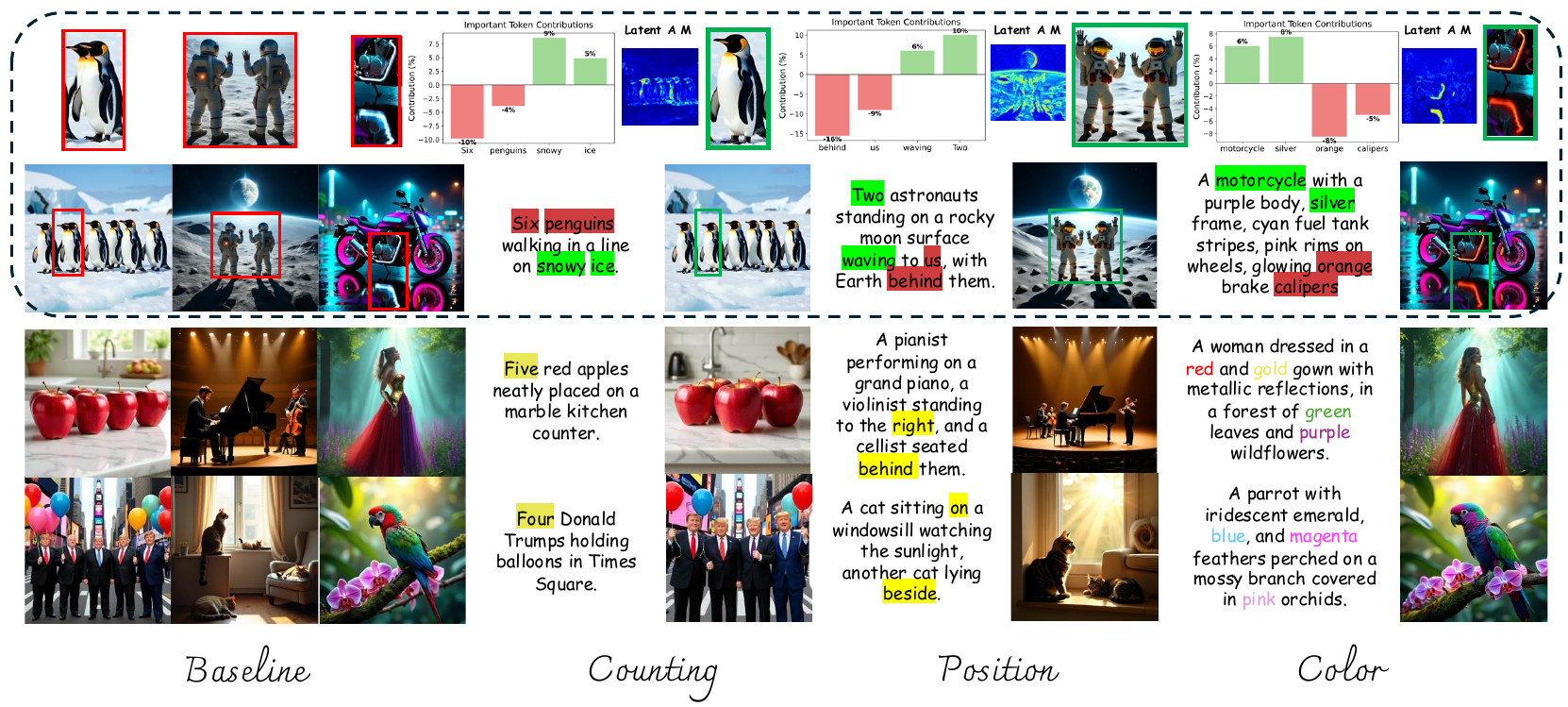}
  \end{minipage}%
  \hfill
  \begin{minipage}[t]{0.28\linewidth}
    \vspace{8pt}
    \includegraphics[width=\linewidth]{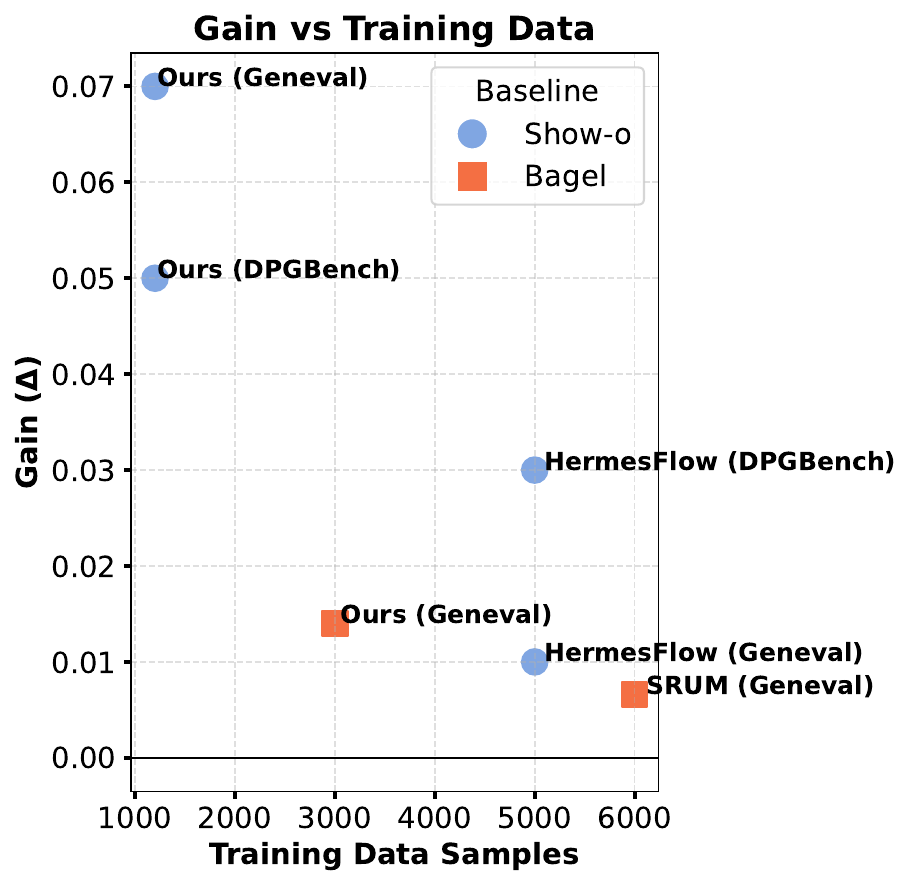}
  \end{minipage}%
}\vspace{-5mm}
    \captionof{figure}{We propose \textbf{xLARD}, a self-correcting generation framework guided by explainable latent rewards. \textbf{Left:} Compared to the baseline, xLARD more faithfully adheres to prompts involving counting, spatial positioning, and color composition. Each example pairs the baseline output with our result for the same prompt. \textbf{Right:} Performance gain versus training data size on Geneval and DPGBench benchmarks, showing that xLARD achieves higher gains with fewer samples.}\vspace{-2mm}
    \label{fig:teaser}
\end{center}
]

\footnotetext[2]{Corresponding author.}

\begin{abstract}
Despite significant progress in text-to-image generation, aligning outputs with complex prompts remains challenging, particularly for fine-grained semantics and spatial relations. This difficulty stems from the feed-forward nature of generation, which requires anticipating alignment without fully understanding the output. In contrast, evaluating generated images is more tractable. Motivated by this asymmetry, we propose \textbf{xLARD}, a self-correcting framework that uses multimodal large language models to guide generation through E\textbf{x}plainable \textbf{LA}tent \textbf{R}ewar\textbf{D}s. xLARD introduces a lightweight corrector that refines latent representations based on structured feedback from model-generated references. A key component is a differentiable mapping from latent edits to interpretable reward signals, enabling continuous latent-level guidance from non-differentiable image-level evaluations. This mechanism allows the model to understand, assess, and correct itself during generation. Experiments across diverse generation and editing tasks show that xLARD improves semantic alignment and visual fidelity while maintaining generative priors. %, offering a data-efficient and generalizable solution to bridging the gap between comprehension and synthesis. 
Code is available at \url{https://yinyiluo.github.io/xLARD/}.

\end{abstract}    
\section{Introduction}
\label{sec:intro}

Recent advances in large multimodal models (LMMs) such as GPT-4V~\citep{yang2023dawnlmmspreliminaryexplorations}, Gemini~\citep{team2023gemini}, Qwen2.5-VL~\citep{bai2025qwen2}, and Bagel~\citep{zhang2025unified} have significantly improved visual-language understanding and generation. These models demonstrate strong capabilities in open-ended visual reasoning~\citep{thawakar2025llamav, bigverdi2025perception}, attribute recognition~\citep{liu2024multi, zou2024eiven}, and compositional understanding~\citep{fu2025evaluating, li2025enhancing, pmlr-v235-li24s}. However, despite their impressive comprehension abilities, they often struggle to faithfully express that understanding during image generation~\citep{hong2025reinforcing, yang2025hermesflow, yan2025can, mao2025unirl, jin2025srum}.

For example, as shown in the count pairs in Figure~\ref{fig:teaser}, when prompted with ``Six penguins walking in a line on snowy ice'', the baseline model (a standard text-to-image model trained with cross-modal supervision) produces an image with incorrect object count and arrangement, despite correctly understanding the prompt. This suggests a core asymmetry: multimodal models can understand correctly but generate incorrectly. We attribute this mismatch to an architecturally unified yet functionally decoupled design between understanding and generation. While the understanding component captures high-level semantics from input modalities, the generator synthesizes outputs in pixel space without explicit access to the model's internal reasoning~\citep{li2023blip, alayrac2022flamingo, chen2025janus, liu2024playground}. Although jointly trained, these components become functionally decoupled at inference time, which often leads to failures in structured reasoning tasks such as spatial grounding or object consistency.

This gap has been addressed through three main paradigms: post-training correction, post-hoc refinement, and training-free methods. Post-training approaches~\citep{xie2025reconstruction, jin2025srum, yang2025hermesflow, xiao2025omnibridge, hong2025reinforcing, yan2025can, mao2025unirl} fine-tune the generator using large-scale feedback, often via reinforcement learning or instruction tuning. While effective, they require heavy supervision, additional data, and expensive retraining, and offer limited interpretability. Post-hoc methods~\citep{chen2025grpocareconsistencyawarereinforcementlearning, ir.2025.13} apply consistency checks or auxiliary models after generation, but provide no control during the process. Training-free approaches~\citep{wu2023selfcorrectingllmcontrolleddiffusionmodels, tewel2024trainingfreeconsistenttexttoimagegeneration, cao2023masactrltuningfreemutualselfattention} bypass retraining entirely, but rely on ad hoc rules or external heuristics, often lacking semantic transparency and model-internal reasoning.

The above limitations motivate our key insight: it is easier and more tractable  to evaluate and then correct the generated images than directly generating faithful contents.
Rather than relying on post-training or post-generation correction, we propose to treat the model's internal comprehension as a real-time guidance signal during generation.
Specifically, we introduce \textbf{xLARD} (E\textbf{x}plainable \textbf{LA}tent \textbf{R}ewar\textbf{D}), a self-correcting framework that integrates the model's own understanding into the generative process through latent-space interventions.
xLARD adds a lightweight residual corrector that refines intermediate latent representations using reward signals derived from interpretable visual-semantic reasoning, including aspects such as counting, color, and spatial alignment. For each prompt, the model first produces a latent representation that is modified by the corrector before being decoded into an image. The resulting image is supervised by a high-quality reference, guiding the corrector to align latents with the intended semantics. The corrector is trained to shift latents toward regions that produce more accurate generations, without altering the backbone.

To enable learning from structured yet non-differentiable feedback, we design a differentiable mapping from latent edits to interpretable reward signals. This allows the model to receive continuous guidance based on how well its generation aligns with the intended meaning. We further adopt a PPO~\citep{schulman2017proximal}-based reinforcement objective, where the reward is obtained from the model's own evaluation of consistency between prompt and image. Because this feedback reflects specific semantic aspects, the corrections are not only effective but also explainable.

As shown in Figure~\ref{fig:teaser}, xLARD improves generation fidelity in object counting, spatial positioning, and color composition. It outperforms or matches post-training methods while requiring significantly less data. The corrector is lightweight, operates during generation, and preserves pretrained generative priors. By leveraging internal model understanding as structured reward feedback, \method enables interpretable correction: important tokens’ contributions are visualized in red (misaligned) and green (aligned) to highlight semantic consistency across color, position, and counting aspects. Together with latent activation maps (LAMs) that localize the model’s focus regions, these visualizations illustrate how semantic understanding drives corrective behavior, offering a general, efficient, and explainable approach for any text-to-image model coupling understanding and generation in latent space.

Our contributions are threefold:
\begin{itemize}
    \item We propose xLARD, a plug-and-play framework for text-to-image generation that performs semantic self-correction in latent space. It integrates a lightweight semantic corrector trained with explainable latent rewards, leveraging the frozen model’s own comprehension to guide multi-aspect corrections, including count, color, and position.
    \item Our approach makes interpretability a core design principle: each correction step is grounded in semantic reasoning and can be decomposed into human-understandable components.
    \item Extensive experiments on diverse generation and editing tasks demonstrate that our method improves semantic alignment and visual fidelity, achieving a \textbf{+4.1\%} gain on \textbf{Geneval} and \textbf{+2.97\%} on \textbf{DPGBench}, while requiring significantly less data and computation than post-training baselines.
\end{itemize}

\section{Related work}
\label{sec:related_work}

\noindent \textbf{Visual Generative Models.}
Recent text-to-image models have made substantial progress using diffusion-based architectures~\citep{ramesh2022hierarchical, saharia2022photorealistictexttoimagediffusionmodels, rombach2022highresolutionimagesynthesislatent, podell2023sdxl}, enabling controllable and high-fidelity image synthesis.
They generate images by progressively denoising latent representations conditioned on text embeddings, producing diverse and realistic outputs. However, they continue to struggle with semantic precision, such as accurate object counts, spatial relationships, and fine-grained attribute alignment, especially under complex or compositional prompts~\citep{liu2024multi, fu2025evaluating, hong2025reinforcing}. While stronger language encoders and large-scale multimodal pretraining have improved overall alignment~\citep{li2023blip, alayrac2022flamingo, yang2023dawnlmmspreliminaryexplorations, bai2025qwen2}, a gap remains between textual understanding and faithful visual realization.

\begin{figure*}[t!]
    \vspace{-.35in}
    \centering
    \includegraphics[width=.95\linewidth]{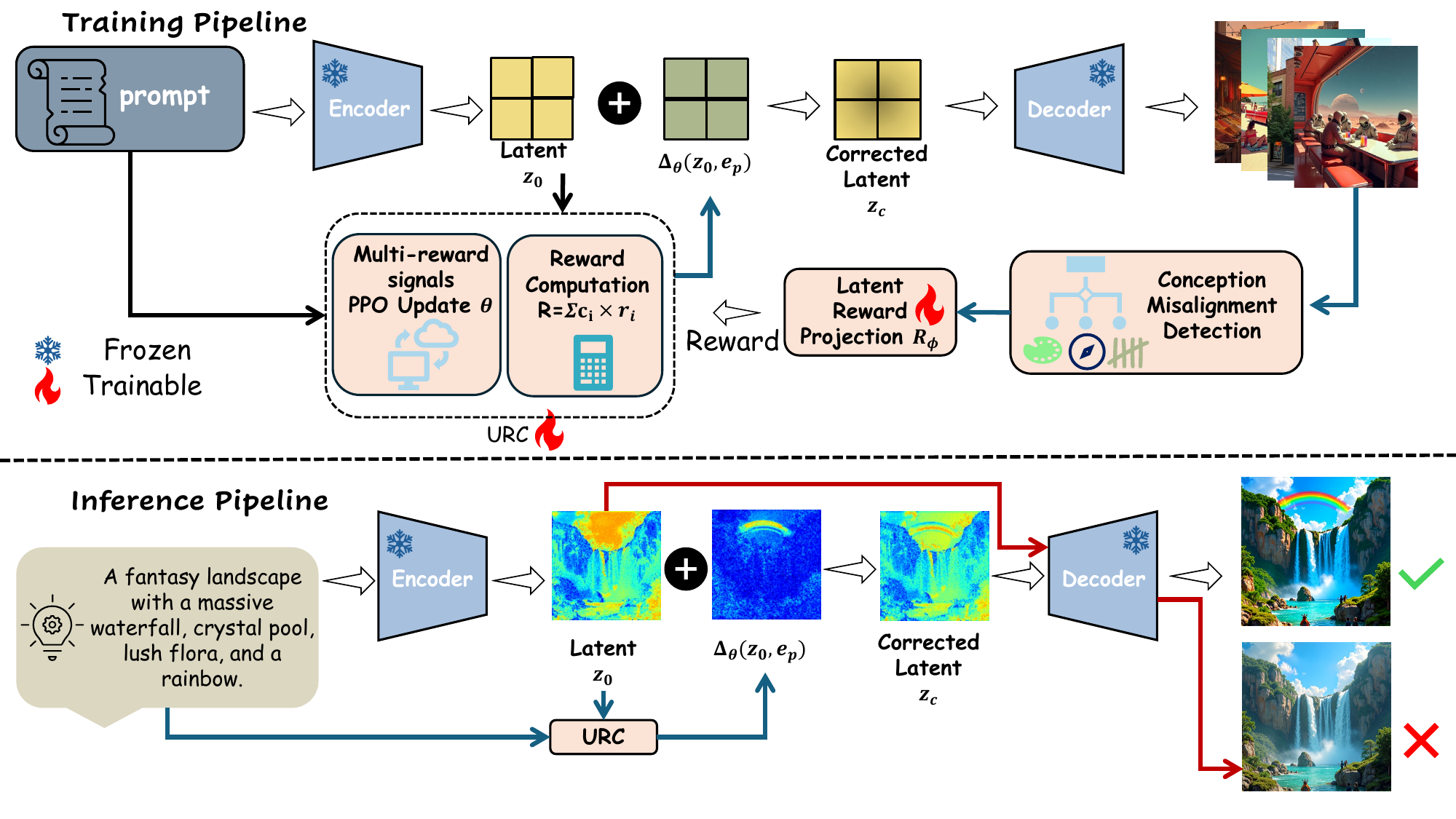}
    \vspace{-.25in}
    \caption{\textbf{Overview of the \method framework.}
    Given a prompt $p$, the frozen backbone encodes it into a latent representation $z_0$.
    The residual corrector $\Delta_\theta$ refines $z_0$ under multi-dimensional reward guidance, producing a corrected latent $z_c$ that is decoded into an image $\hat{x}$.
    Image-level rewards are projected back to the latent space via a learnable reward projector $R_\phi$, allowing end-to-end, interpretable correction learning.
    During inference, URC functions as a lightweight latent modifier with no additional sampling or retraining.}
    \label{fig:residual_pipeline}
    \vspace{-.20in}
\end{figure*}

\noindent \textbf{Semantic Alignment and Latent Refinement.}
To bridge this gap, several approaches refine latent representations to improve prompt adherence. CLIP-guided optimization~\citep{gal2022imageworthwordpersonalizing, patashnik2021stylecliptextdrivenmanipulationstylegan} and classifier-free guidance~\citep{dhariwal2021diffusionmodelsbeatgans, ho2022classifierfreediffusionguidance} steer generation toward text-consistent outputs, but often degrade visual quality or introduce instability. Other methods incorporate residual adapters or fine-tune the backbone to improve alignment~\citep{xiao2025omnibridge, jin2025srum, yang2025hermesflow}, though these typically require large-scale retraining or additional supervision. Our work differs by introducing a lightweight semantic corrector that refines latent features during generation, operating in a plug-and-play fashion without modifying the backbone. The corrector is trained end-to-end using structured reward signals derived from the model's own multimodal understanding.

\noindent \textbf{Self-Correction and Understanding-Driven Control.}
Several training-free strategies have explored inference-time correction using self-attention mechanisms or mutual feedback~\citep{wu2023selfcorrectingllmcontrolleddiffusionmodels, tewel2024trainingfreeconsistenttexttoimagegeneration, cao2023masactrltuningfreemutualselfattention}. While efficient, these approaches often depend on handcrafted heuristics or auxiliary modules, limiting generalizability and interpretability. Other works leverage multimodal models for post-hoc reward scoring or CLIP-based consistency evaluation~\citep{hong2025reinforcing, chen2025grpocareconsistencyawarereinforcementlearning}, but apply only after image generation, offering no real-time correction. In contrast, our method integrates multimodal understanding directly into the generative loop. We train a latent semantic corrector using understanding-guided reinforcement, where interpretable reward signals guide latent refinement as generation unfolds. This enables real-time, interpretable, and model-agnostic self-correction grounded in the model's own reasoning, without requiring retraining or external discriminators.

\section{Method}
\label{sec:method}

We introduce \textbf{xLARD}, a general framework that improves text-to-image generation through interpretable latent-space reinforcement.  
As illustrated in \Cref{fig:residual_pipeline}, xLARD operates as a self-correcting loop that integrates the model’s own multimodal understanding into the generative process.  
It consists of three key components:

\begin{enumerate}
    \item \textbf{Understanding-Guided Reinforcement Corrector (URC)} ($\Delta_\theta$): a policy network that refines latent representations through residual updates guided by semantic rewards.
    \item \textbf{Conception Misalignment Detection Module (CMD)}: a module that detects and quantifies image–prompt inconsistencies, providing image-level guidance to the reward module.
    \item \textbf{Explainable Latent Reward Projection Module} ($R_\phi$): a differentiable reward projector that maps latent activations to interpretable semantic feedback across count, color, and position dimensions.
\end{enumerate}

Together, these components enable the model to evaluate and correct its own generations in real time (without retraining, additional supervision, or backbone modification).  
We next describe each module in detail.

\subsection{Reinforcement Corrector}

As shown in \Cref{fig:residual_pipeline}, given a pretrained text-to-image generator $\mathcal{M}$ with encoder–decoder structure $(\mathcal{E}, \mathcal{D})$, URC inserts a corrector $\Delta_\theta$ in the latent space.
For a text prompt $p$, the encoder $\mathcal{E}$ produces a latent code $z_0 = \mathcal{E}(p)$.
The corrector then applies a small understanding-guided shift:
\begin{equation}
    z_c = z_0 + \alpha \cdot \Delta_\theta(z_0, e_p),
\end{equation}
where $e_p$ is the prompt embedding and $\alpha$ controls residual strength.
The decoder $\mathcal{D}$ generates an image $\hat{x} = \mathcal{D}(z_c)$.

\noindent \textbf{Training within a Frozen Pipeline.}
URC learns without modifying the backbone.
Given a prompt–reference pair $(p, x^*)$, the generated image $\hat{x}$ receives an image-level reward that captures alignment, realism, and attribute correctness.
This reward is then projected back into latent space through a differentiable reward projector $R_\phi$, learning a continuous mapping:
\begin{equation}
r_\text{latent} = R_\phi(z_c, e_p) \approx r_\text{image}(\hat{x}, p, x^*).
\end{equation}
The corrector $\Delta_\theta$ is optimized end-to-end using $r_\text{latent}$, allowing gradient-based updates even when the original reward is non-differentiable.

\noindent \textbf{Inference.}
At inference, URC simply applies $\Delta_\theta$ on a single latent without reward computation or additional sampling, functioning as an efficient latent-level modifier transferable to various diffusion or VAE-based architectures.

\subsection{Conception Misalignment Detection}

While URC provides localized latent corrections, it requires reliable high-level guidance on whether the generated image semantically aligns with the intended prompt.  
The \textbf{Conception Misalignment Detection Module (CMD)} fulfills this role by identifying image-level mismatches.  
CMD thus acts as a semantic evaluator that ensures URC’s residual updates remain globally consistent with the user’s intent.

\subsubsection{Task-Specific Rewards}
Prior models frequently misrepresent object quantities, miscolor entities, or misplace objects relative to one another, even when the prompt is correctly understood. To explicitly link the model’s internal understanding to observable visual correctness, we design interpretable task-specific sub-rewards along three orthogonal dimensions: counting, color, and position. These dimensions together aim to address these common failure modes. Each sub-reward is computed directly from the backbone’s feature maps and the prompt’s linguistic structure, allowing URC to quantify how well the model’s latent representation satisfies the underlying textual intent.

\textbf{Counting Reward.}
We extract token-level attention maps from the image encoder’s feature maps and identify the activation regions corresponding to each object token (e.g., ``dog'', ``apple'').
Let $A_t(h,w)$ denote the attention activation for token $t$ at spatial position $(h,w)$.
The number of distinct activation clusters is estimated via connected-component analysis on $A_t$, giving the predicted count $\hat{n}_t$.
We parse the prompt to obtain the target count $n_t$ (e.g., ``two dogs'' $\Rightarrow n_{\text{dog}} = 2$).
The reward encourages numerical consistency:
\begin{equation}
r_\text{count} = \exp\!\left(-\frac{|\hat{n}_t - n_t|}{n_t}\right),
\end{equation}
which softly penalizes over- or under-counting while remaining differentiable.

\textbf{Color Reward.}
We extract the set of color-related words $\mathcal{C} = \{\text{red}, \text{blue}, \text{green}, \dots\}$ from the prompt and compute their text embeddings $\{e_c\}$ using model's text encoder.
Given the patch-level image features $\{f_i\}$ from the backbone, we compute patch--color similarities $s_{i,c} = \cos(f_i, e_c)$.
For each color word $c$, the color reward is defined as:
\begin{equation}
r_\text{color} = \frac{1}{|\mathcal{C}|} \sum_{c \in \mathcal{C}} \max_i s_{i,c},
\end{equation}
which measures how strongly each color concept is expressed in any patch of the generated image.
This reward encourages precise attribute realization and disentangles color fidelity from other factors such as texture or shape.

\textbf{Position Reward.}
Spatial relation words (e.g., ``left of'', ``right of'', ``on top of'', ``under'') are parsed from the prompt to form a set of positional constraints $\mathcal{R}$.
For each relation $(t_a, t_b, r) \in \mathcal{R}$, we locate the entity centers $p_a, p_b$ on the image encoder’s activation map via attention-weighted centroids of their corresponding token maps:
\begin{equation}
p_t = \frac{\sum_{h,w} (h,w) \cdot A_t(h,w)}{\sum_{h,w} A_t(h,w)}.
\end{equation}
We then compute the directional consistency between the predicted and target spatial relations using a differentiable indicator:
\begin{equation}
r_\text{pos} = \frac{1}{|\mathcal{R}|} \sum_{(a,b,r) \in \mathcal{R}}
\sigma\!\left( \frac{(p_b - p_a) \cdot v_r}{\tau} \right),
\end{equation}
where $v_r$ is the canonical direction vector for relation $r$ (e.g., ``left of'' $\Rightarrow v_r = [-1, 0]$), $\tau$ controls smoothness, and $\sigma$ denotes the sigmoid function.
This yields a continuous positional reward that aligns geometric reasoning in latent space with textual relational understanding.

\textbf{Joint Task Reward.}
The total task-specific reward combines the three interpretable signals:
\begin{equation}
r_\text{task} = \lambda_\text{count} r_\text{count} +
\lambda_\text{color} r_\text{color} +
\lambda_\text{pos} r_\text{pos},
\end{equation}
where the $\lambda$ terms are not hyperparameters to be tuned, but are dynamically modulated by the confidence head based on the model’s uncertainty for each task aspect.

\subsection{Latent Reward Projection}
Direct backpropagation from image-level reward is often infeasible because the decoding process is non-differentiable.
\method addresses this by introducing a learnable \textbf{latent reward projector} $R_\phi$, trained to approximate image-level feedback using the latent activations and prompt embedding:
\begin{equation}
r_\text{latent} = R_\phi(z_c, e_p) \in \mathbb{R}^3,
\end{equation}
corresponding to the three interpretable sub-rewards above.
Once trained, $R_\phi$ provides differentiable reward gradients to $\Delta_\theta$, allowing reinforcement updates purely within latent space.

\noindent \textbf{Policy Optimization.}
The corrector is optimized to maximize the expected latent reward:
\begin{equation}
\theta^* = \arg\max_\theta \mathbb{E}_{p \sim \mathcal{P}} [R_\phi(z_0 + \Delta_\theta(z_0, e_p), e_p)].
\end{equation}
We adopt Proximal Policy Optimization (PPO)\citep{schulman2017proximal} for stable updates, combining stochastic exploration with reward-weighted gradients:
\[
\nabla_\theta \mathcal{L} = - (R_\phi - b) \nabla_\theta \log \pi_\theta(\Delta_\theta | z_0, e_p),
\]
where $b$ is a learned baseline reducing variance.

\subsection{Intrinsic Interpretability}

URC achieves intrinsic interpretability through three mechanisms:
\begin{enumerate}
    \item \textbf{Decomposed Reward Dimensions.}
   Each sub-reward (counting, position, color) corresponds to a well-defined latent behavior, making the learned residuals directly explainable.
   \item \textbf{Latent Activation Maps (LAM).}
   The magnitude of $\Delta_\theta$ reveals where corrections are concentrated:
   \begin{equation}
   \text{LAM}(h,w) = \sum_c |\Delta_\theta(z_0, e_p)[c, h, w]|
   \vspace{-2mm}
   \end{equation}
   Correlating LAM with token-level attention yields token-to-region explanations of what the model corrected and why (details in \Cref{fig:latent_cam}, \Cref{fig:token_contrib}).
   \item \textbf{Latent Reward Projection.}
   Rather than relying solely on gradient-based parameter updates, URC promotes latent-level understanding by explicitly associating reward signals with interpretable latent dimensions.  
   The reward projector $R_\phi$ translates latent activations into semantic components such as “object count,” “color,” or “spatial position,” enabling the model to reason about how each factor influences generation quality.  
   This design provides a transparent view of the correction process, clarifying both the motivation and the effect of each latent modification.
\end{enumerate}

\noindent \textbf{Mechanistic Insight.}
URC enhances generation through a closed-loop, self-understanding mechanism rather than opaque fine-tuning.
By learning a differentiable mapping from latent corrections to interpretable rewards, it converts non-differentiable image-level signals into latent-level guidance.
This allows the model to \emph{understand, evaluate, and correct itself} in a continuous and explainable manner, bridging the gap between comprehension and generation.

\section{Experiment}

We conduct extensive experiments to evaluate the effectiveness of \method across a range of image generation benchmarks. Our goal is to demonstrate that the proposed approach enhances semantic fidelity, compositional understanding, and overall image quality compared to state-of-the-art (SOTA) baselines (more details can be found in the supplementary material).

\subsection{Evaluation on Text-to-Image Generation}
\label{sec:benchmark}

\begin{figure*}[h]
    \vspace{-.3in}
    \centering
    % Top image
    \includegraphics[width=\linewidth]{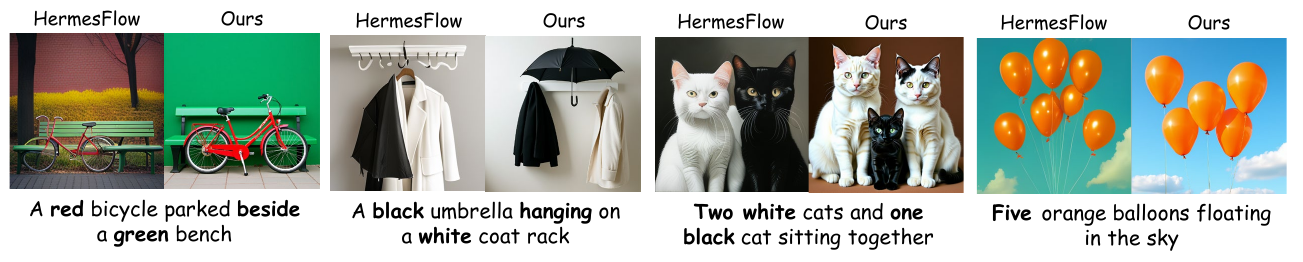}
    \vspace{0.5em}
    \small{(a) Image Generation}
    
    \vspace{-.08in}
    
    % Bottom image
    \includegraphics[width=\linewidth]{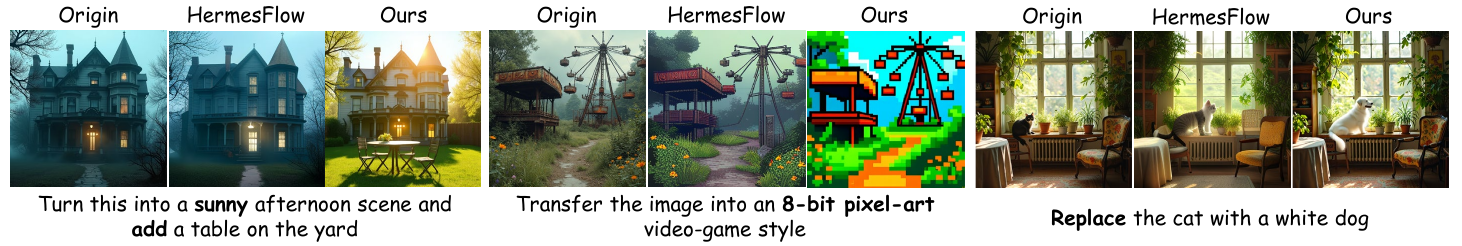}
    \vspace{0.5em}
    \small{(b) Image Editing}

    \vspace{-.1in}
    \caption{Qualitative comparison of image generation/editing performance between HermesFlow and our proposed approach.}
    \label{fig:image_comparison}
    \vspace{-.15in}
\end{figure*}

\begin{table}[t]
\centering
\renewcommand{\arraystretch}{1.2} % 控制行高
\setlength{\tabcolsep}{1pt} % 缩小列间距
\vspace{1mm}

\caption{Comparison of generative and editing performance across benchmarks. \\
\footnotesize{\textsuperscript{*}: Results are re-generated using the official pre-trained models.}}
\label{tab:benchmark_results}

\vspace{-.1in}
\hspace*{-2mm}
{\footnotesize
\begin{tabular}{@{\hspace{0pt}}l
p{1.6cm}<{\centering}p{1.2cm}<{\centering}
p{1.4cm}<{\centering}
p{1.4cm}<{\centering}@{\hspace{0pt}}}
\toprule
Method & Type & Params & DPG-Bench & GenEval  \\

\midrule
OmniGen & AR+Diffusion & 7B & 81.16 & 0.68 \\
Show-O & AR+Diffusion & 1.3B & 67.27 & 0.68 \\
BAGEL\textsuperscript{*} & AR+RAG & 14B & 84.07 & 0.79 \\
FLUX-dev & Diffusion & 12B & 84.00 & 0.68 \\
UniWorld-V1 & AR+Diffusion & 20B & 81.38 & 0.80 \\
Emu3 & AR & 8B & 80.60 & 0.54 \\
Janus-pro & AR & 7B & 84.19 & 0.80 \\
OmniGen2\textsuperscript{*} & Diffusion+AR & 7B & 83.48 & 0.77 \\
\rowcolor{blue!8}\textbf{\method} & \cellcolor{blue!8} & \cellcolor{blue!8}\textbf{} & \cellcolor{blue!8}\textbf{86.45} & \cellcolor{blue!8}\textbf{0.81} \\
\bottomrule
\end{tabular}
}
\vspace{-.25in}
\end{table}

We assess the performance of our method on several standard T2I benchmarks designed to measure both low-level and high-level alignment between textual prompts and generated images. The evaluation emphasizes compositional reasoning, object fidelity, and attribute accuracy, which are crucial indicators of a model’s semantic grounding. We compare \method against several popular approaches, including diffusion-based models (Omnigen \citep{xiao2025omnigen}, OmniGen2 \citep{wu2025omnigen2}, Show-O \citep{xie2024show}, and UniWorld-V1 \citep{lin2025uniworld}), retrieval-augmented models (BAGEL \citep{deng2025emerging}), and autoregressive large language models (Janus-pro \citep{chen2025janus}, GPT-4o-Image \citep{openai2024gpt4o}, and Emu3 \citep{wang2024emu3}). 

We employ two major benchmark suites to comprehensively evaluate model performance. The first, \textbf{GenEval} \citep{ghosh2023geneval}, focuses on compositional understanding, including the accuracy of generated objects, spatial relations, and attributes such as color and count. It provides a quantitative measure of a model’s ability to synthesize semantically correct and visually coherent scenes.  
The second, \textbf{DPG-Bench} \citep{liu2025step1x}, evaluates models from a linguistic–visual alignment perspective, categorizing performance into five L1 categories: \textit{entity}, \textit{attribute}, \textit{relation}, \textit{global}, and \textit{other}. These categories capture both fine-grained object details and broader scene consistency, allowing a holistic understanding of model behavior.

As shown in Table~\ref{tab:benchmark_results}, our approach achieves superior or comparable performance across all benchmarks. Specifically, \method improves the GenEval score by a notable margin, indicating stronger compositional reasoning, and maintains competitive accuracy on DPG-Bench, reflecting enhanced cross-modal understanding and alignment. The visual comparison in Figure~\ref{fig:image_comparison}a further supports these findings, our method produces scenes that more faithfully reflect textual semantics, with more coherent spatial composition and color correspondence.

\vspace{-.08in}
\subsubsection{Fine-Grained Benchmark Analysis}
\label{sec}

To better understand the improvements, we conduct category-level analyses on both GenEval and DPG-Bench.

\noindent \textbf{GenEval Category Analysis.}
Our method consistently outperforms the baseline across all sub-metrics as shown in \Cref{tab:geneval_combined}, with the most notable gains in counting (+9.4\%) and color/attribute binding. These improvements highlight the model’s enhanced capability to understand and maintain fine-grained correspondences between textual concepts and visual elements, such as object quantities, spatial placement, and color associations.

\begin{table}[t]
\centering
\caption{GenEval detailed evaluation metrics (\%) across different baselines and backbones.}
\label{tab:geneval_combined}
\vspace{-0.1in}
\setlength{\tabcolsep}{.4pt}
%\hspace*{-6mm}
\resizebox{\linewidth}{!}{
{\footnotesize
\begin{tabular}{@{}lccccccc@{}}
\toprule
Method & Single Obj & Two Obj & Cnt & Colors & Position & Color Attr & Overall \\
\midrule
% \multicolumn{8}{l}{\textit{Diffusion-based Models}} \\
OmniGen2 & 100 & 92.93 & 69.12 & 85.88 & 45.52 & 68.75 & 77.03 \\
\rowcolor{blue!8}OmniGen2 + Ours & \cellcolor{blue!8}100 & \cellcolor{blue!8}93.69 & \cellcolor{blue!8}78.44 & \cellcolor{blue!8}92.11 & \cellcolor{blue!8}48.75 & \cellcolor{blue!8}73.75 & \cellcolor{blue!8}81.29 \\ \midrule

\addlinespace[2pt]
% \multicolumn{8}{l}{\textit{Retrieval-augmented Models}} \\
Bagel & 99.12 & 92.34 & 77.45 & 88.21 & 50.89 & 63.78 & 78.97 \\
\rowcolor{blue!8}Bagel + Ours & \cellcolor{blue!8}99.85 & 93.95 & 80.55 & 87.33 & 60.72 & 65.89 & 81.38 \\ \midrule

\addlinespace[2pt]
% \multicolumn{8}{l}{\textit{Autoregressive Models}} \\
Show-O & 98.21 & 80.35 & 65.89 & 84.12 & 30.78 & 49.95 & 68.05 \\
\rowcolor{blue!8}Show-O + Ours & 98.95 & 90.88 & 74.56 & 85.72 & 41.87 & 55.34 & 74.89 \\
\bottomrule
\end{tabular}
}
}
\vspace{-.25in}
\end{table}

\noindent \textbf{DPG-Bench Category Analysis.}
DPG-Bench provides a complementary evaluation from a multimodal reasoning perspective, as shown in \Cref{tab:dpgbench_combined}. Our model achieves balanced improvements across all Level-1 categories, with the largest margins in the entity and attribute dimensions. This indicates a stronger understanding of inter-object relationships and attribute grounding, which are typically difficult for diffusion-based generators to model reliably.

\vspace{-.05in}
\subsubsection{Cross-Backbone Evaluation}

\label{sec:backbone}

To assess generality, we integrate \method into different backbones. The consistent gains reported in \Cref{tab:geneval_combined,tab:dpgbench_combined} demonstrate the plug-and-play nature of our method, highlighting its robustness and adaptability to different T2I architectures.

\begin{table}[t]
\centering
\caption{DPG-Bench detailed evaluation metrics (\%) across different baselines and backbones.}
\label{tab:dpgbench_combined}
\vspace{-0.1in}
\setlength{\tabcolsep}{2pt}
\hspace*{-1mm}
{\footnotesize
\begin{tabular}{@{}lcccccc@{}}
\toprule
Method & Global & Entity & Attribute & Relation & Other & Overall \\
\midrule
% \multicolumn{7}{l}{\textit{Diffusion-based Models}} \\
OmniGen2 & 82.07 & 90.22 & 87.82 & 93.26 & 82.26 & 83.48 \\
\rowcolor{blue!8}OmniGen2 + Ours & \cellcolor{blue!8}82.76 & \cellcolor{blue!8}92.41 & \cellcolor{blue!8}89.90 & \cellcolor{blue!8}94.07 & \cellcolor{blue!8}86.60 & \cellcolor{blue!8}86.45 \\ \midrule

\addlinespace[2pt]
% \multicolumn{7}{l}{\textit{Retrieval-augmented Models}} \\
Bagel & 80.46 & 90.28 & 87.48 & 92.42 & 83.40 & 84.07 \\
\rowcolor{blue!8}Bagel+Ours & 81.16 & 91.57 & 88.84 & 93.93 & 85.60 & 85.50 \\ \midrule

\addlinespace[2pt]
% \multicolumn{7}{l}{\textit{Autoregressive Models}} \\
Show-O & 79.33 & 75.44 & 78.02 & 84.45 & 60.80 & 67.27 \\
\rowcolor{blue!8}Show-O+Ours & 85.11 & 81.69 & 84.17 & 90.44 & 67.21 & 72.92 \\
\bottomrule
\end{tabular}
}
\vspace{-.1in}
\end{table}

\vspace{-.08in}
\subsection{Evaluation on Image Editing Task}
\label{sec:imgedit}

We evaluate our method on ImgEdit\citep{ye2025imgedit} and GEdit\citep{liu2025step1x} to assess its ability to perform targeted modifications while preserving irrelevant content. As reported in Table~\ref{tab:edit_table}, our approach achieves higher overall scores compared to OmniGen2, indicating improved semantic understanding and finer control over the editing process.  

Figure~\ref{fig:image_comparison}b presents qualitative results, showing that our method produces edits that better preserve semantic fidelity, maintain alignment with the intended modifications, and generate visually coherent and realistic outputs.

% \begin{table}[t]
% \centering
% \caption{ImgEdit}
% \label{tab:imgedit}
% \vspace{-.1in}
% % 调整列间距
% \setlength{\tabcolsep}{2pt}
% \hspace*{-1mm}

% {\footnotesize % 可改为 \scriptsize 或 \small
% \begin{tabular}{@{}lcccccccccc@{}}
% \toprule
% Method & Rep. & Bg. & Style & Adj. & Ext. & Rem. & Rep. & Add & Comp. & Act.  \\
% \midrule
% OmniGen2 & 2.82 & 3.5 & 3.66 & 3.15 & 3.03 & 2.27 & 2.82 & 3.5 &  2.43 & 3.87 \\
% Ours &  3.04 & 3.62 & 3.57 & 3.2 & 3.17 & 2.69 & 3.04 & 3.62 & 2.36 & 4.22  \\
% \bottomrule
% \end{tabular}
% }
% \vspace{-.2in}
% \end{table}

\begin{table}[t]
    \centering
    \caption{Comparison of \textbf{ImgEdit} and \textbf{GEdit} performance.}\vspace{-3mm}
    \label{tab:edit_table}
    \setlength{\tabcolsep}{0.4pt}
    %\hspace*{-6mm}
    \resizebox{.48\textwidth}{!}{
    {\footnotesize
    \begin{tabular}{@{}lccccccccccccc@{}}
    \toprule
     & \multicolumn{10}{c}{\textbf{ImgEdit}} & \multicolumn{3}{c}{\textbf{GEdit}} \\ 
    \cmidrule(lr){2-11} \cmidrule(lr){12-14}
    Method & Rep. & Bg. & Style & Adj. & Ext. & Rem. & Rep. & Add & Cmp. & Act. & SC & PQ & Overall \\
    \midrule
    OmniGen2 & 2.82 & 3.50 & 3.66 & 3.15 & 3.03 & 2.27 & 2.82 & 3.50 & 2.43 & 3.87 & 5.05 & 5.53 & 4.46 \\
    \rowcolor{blue!8}Ours & 3.04 & 3.62 & 3.57 & 3.20 & 3.17 & 2.69 & 3.04 & 3.62 & 2.36 & 4.22 & 5.38 & 6.77 & 5.52 \\
    \bottomrule
    \end{tabular}
    }
    }
    \vspace{-.2in}
\end{table}

\subsection{Interpretability}
\label{sec:interpretability}

To better understand the mechanisms by which our corrector improves generative performance, we perform an interpretability analysis on both the latent space and text-to-latent interactions.

\begin{figure}[t]
    \centering
    \includegraphics[width=\linewidth]{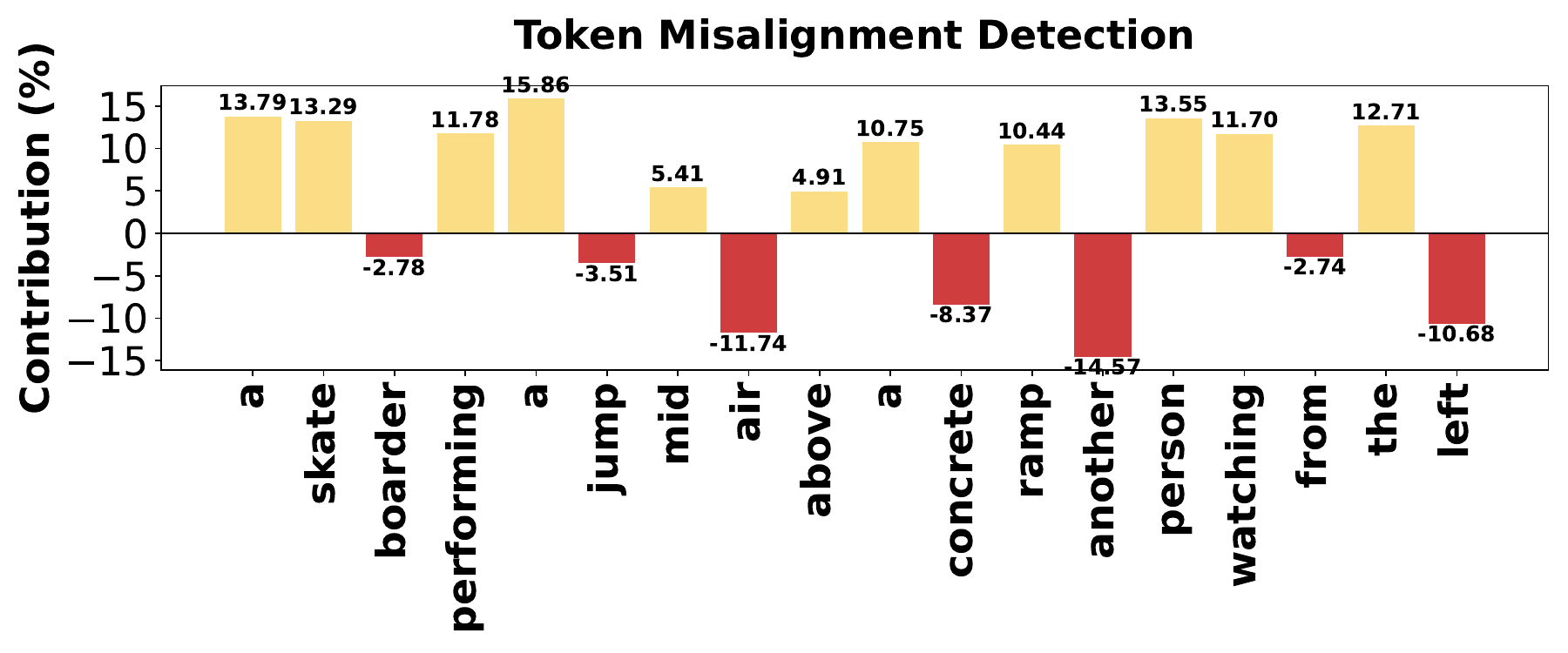} 
    \vspace{-.3in}
    \caption{Token-level contributions for misalignment detection. Positive bars indicate tokens aligned with the image, negative bars indicate tokens driving residual corrections.}
    \label{fig:token_contrib}
    \vspace{-3mm}
\end{figure}

\begin{figure}[t]
    \centering
    \begin{subfigure}[b]{0.32\linewidth}
        \centering
        \includegraphics[width=\linewidth]{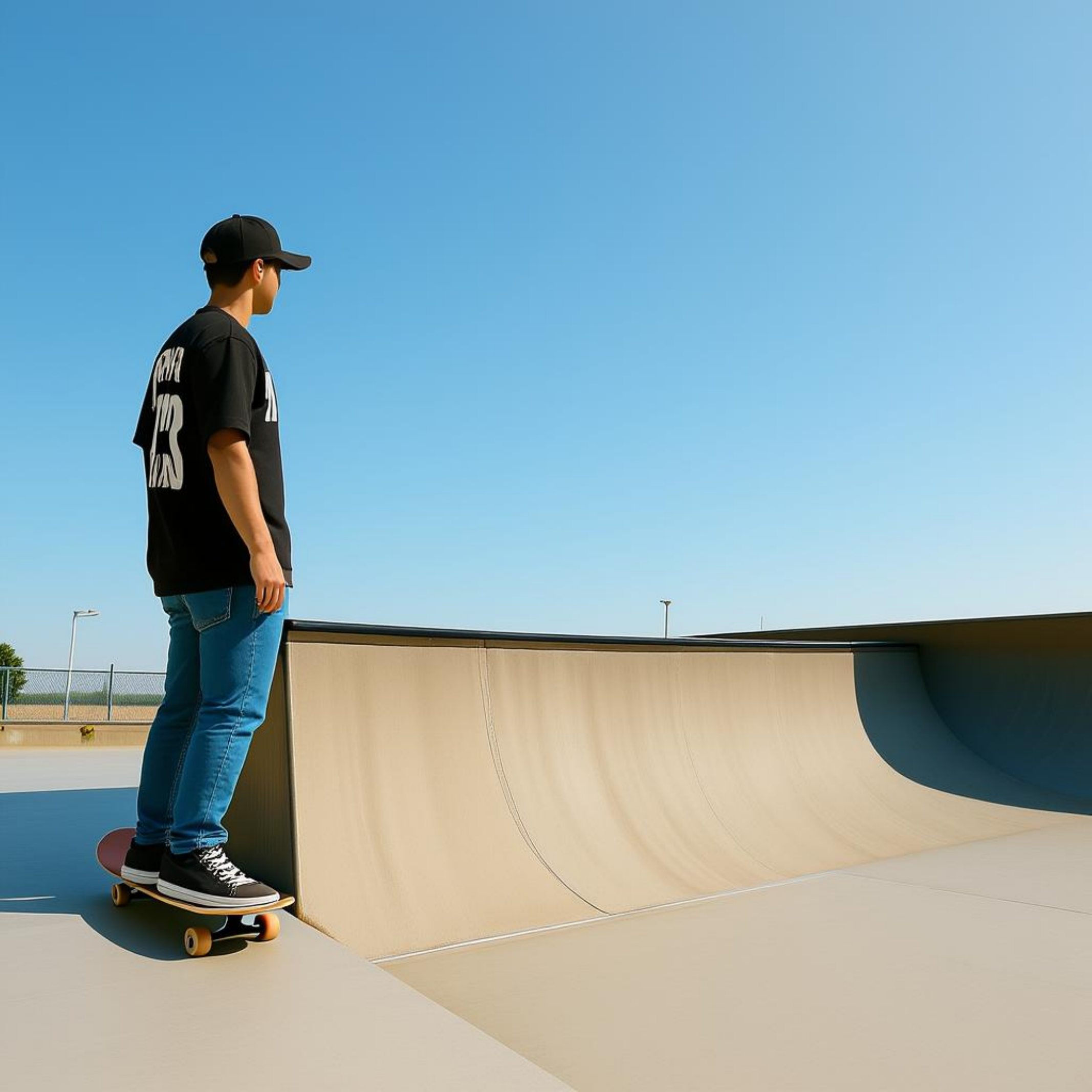}
        \caption{Original image}
    \end{subfigure}
    \hfill
    \begin{subfigure}[b]{0.32\linewidth}
        \centering
        \includegraphics[width=\linewidth]{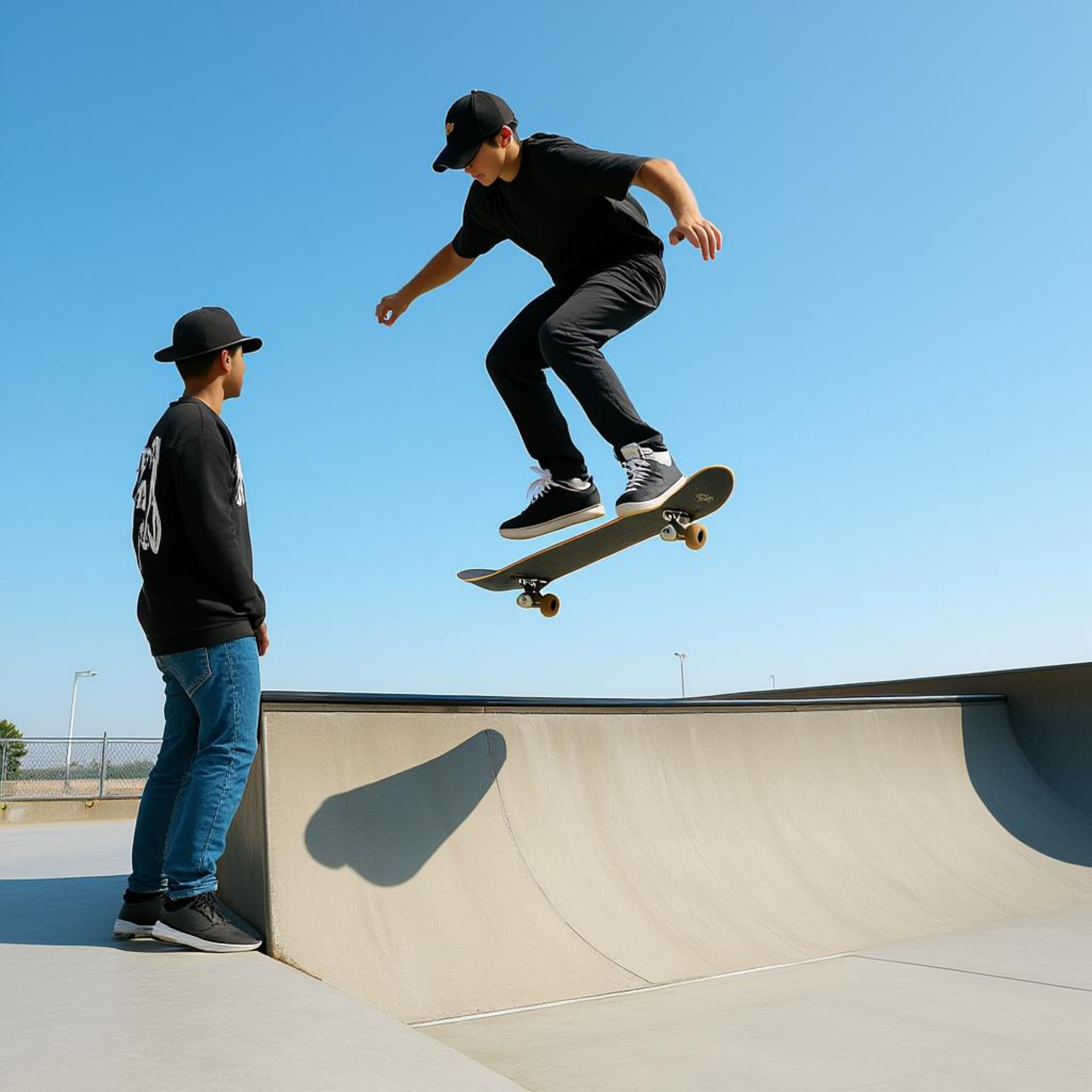}
        \caption{Corrected image}
    \end{subfigure}
    \hfill
    \begin{subfigure}[b]{0.32\linewidth}
        \centering
        \includegraphics[width=\linewidth]{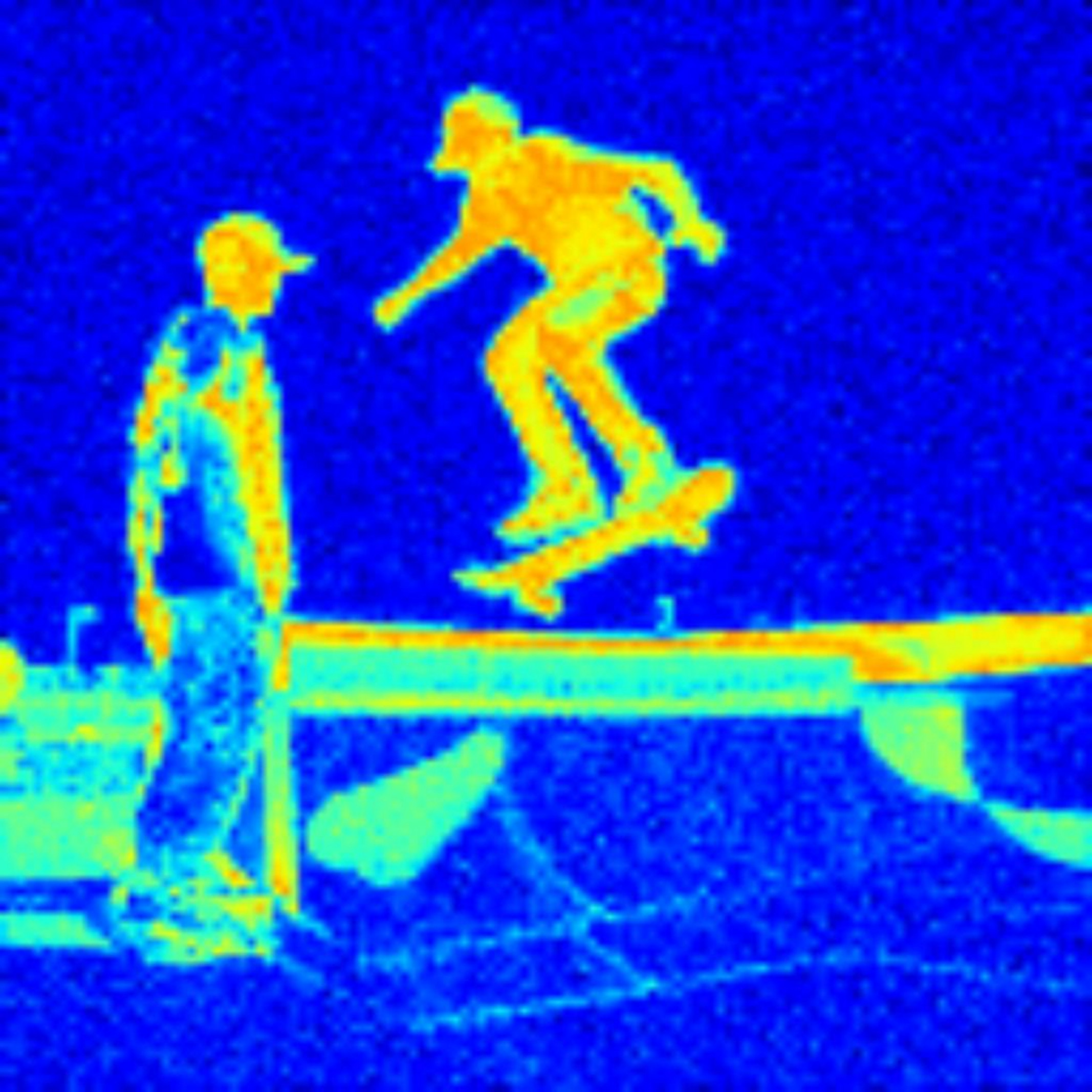}
        \caption{Correction map}
    \end{subfigure}
    \vspace{-3mm}
    \caption{Visualization of latent residual corrections. The high-intensity regions in the correction map indicate where the residual module most strongly adjusts the latent features. The prompt used for this example is \textit{``A skateboarder performing a jump mid-air above a concrete ramp, another person watching from the left.''}}
    \label{fig:latent_cam}
    \vspace{-.2in}
\end{figure}

\begin{figure*}[h]
    \vspace{-.3in}
    \centering
    \includegraphics[width=\linewidth]{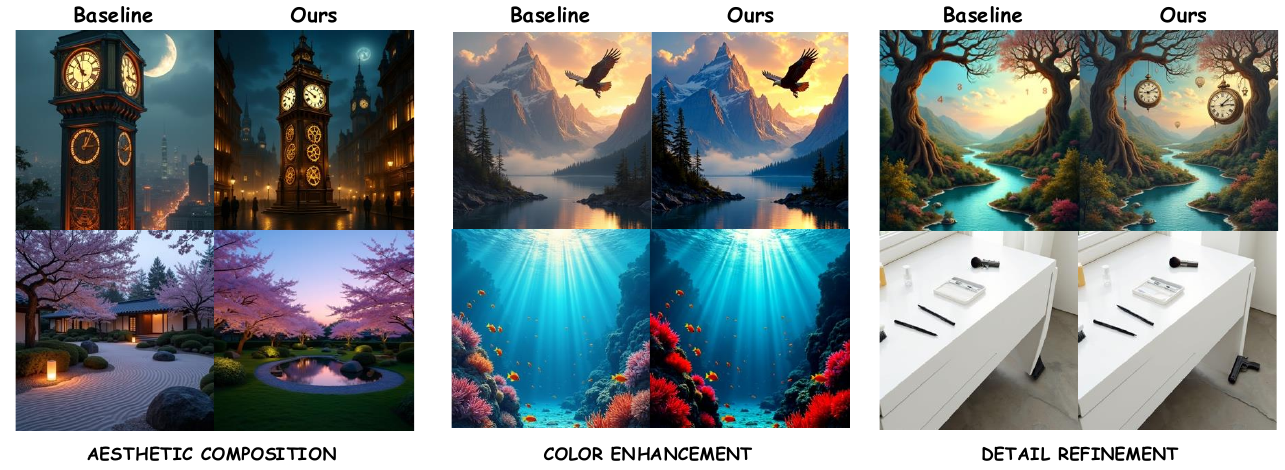}
    \vspace{-7mm}\caption{Illustration of improvements introduced by \method. From left to right: (1) Aesthetic composition. Objects are placed according to the prompt to produce visually coherent layouts; (2) Color enhancement. Colors are adjusted to better match the described scene; (3) Detail refinement. Small details such as textures and secondary objects are corrected for higher fidelity.}
    \label{fig:comparison}
    \vspace{-.20in}
\end{figure*}

\noindent \textbf{Latent Activation Maps (LAM).}
We visualize the residual latent corrections using Latent Activation Maps (LAMs) computed from the latent $\Delta_\theta$ produced by the residual corrector. 
Figure~\ref{fig:latent_cam} shows an example of residual latent corrections. 
These visualizations indicate that our residual corrector focuses on semantically important regions of the image, which helps improve alignment with the textual prompt and corrects physically inconsistent details.

\noindent \textbf{Token Misalignment Detection.}
We analyze token-level contributions to latent corrections to identify where the generated image initially misaligns with the textual prompt. 
Each token's contribution reflects how much the residual corrector modifies the latent representation to reduce this generation-prompt misalignment. 
Figure~\ref{fig:token_contrib} shows an example of normalized token contributions for a given prompt.

\noindent \textbf{Explanation of corrections:}
\begin{itemize}
    \item Token ``skateboarder" contributed strongly: Correction lifts the skateboarder into mid-air, matching the action described in the prompt.
    \item Token ``jump" contributed strongly: Ensures the skateboarder performs a jump above the ramp.
    \item Token ``another person" contributed strongly: Corrects the placement of the second person, removing the skateboard from them to match the prompt.
    \item Token ``ramp" contributed moderately: Refines ramp positioning and alignment for accurate scene composition.
\end{itemize}

Overall, this analysis highlights how the residual corrector systematically identifies and rectifies areas of generation-prompt misalignment, using semantic cues from the text to guide latent modifications. Rather than serving purely as an interpretability tool, token contributions provide insight into which parts of the prompt are most responsible for initial deviations in the generated image and how the model corrects them.

\subsubsection{Quantitative Validation of Interpretability}
\label{sec:interpretability-validation}

We quantitatively evaluate the faithfulness and consistency of xLARD’s interpretability signals by measuring how influential regions and tokens relate to performance gains:
\begin{itemize}
    \item \textbf{Spatial Faithfulness.} Masking high-activation regions in the latent activation map (LAM) and re-decoding images leads to a $6.3\%$ drop in CLIPScore and a $3.8\%$ drop in GenEval, confirming that highlighted regions are causally linked to improved semantic fidelity.

    \item \textbf{Token Contribution vs. Reward Gain.} 
Spearman correlation between per-token contribution magnitudes and semantic alignment reward increases is $\rho=0.71$, indicating that higher-weighted tokens consistently yield larger reward improvements.

    \item \textbf{Cross-Prompt Consistency.} Top-$k$ contributing tokens remain stable across semantically similar prompts (average Jaccard similarity $0.68$), showing coherent token-level explanations under minor prompt variations.
\end{itemize}

These results confirm that xLARD’s interpretability signals faithfully reflect correction behavior and semantic influence rather than being visual artifacts.

\subsection{Ablation Study}
\label{sec:ablation}

To assess the contribution of each component in our framework, we perform ablation studies targeting three key modules: the reinforcement learning (RL) objective, confidence-guided latent modulation, and the latent anchor mechanism. As shown in Table~\ref{tab:ablation_results}, removing the RL component results in a noticeable degradation on both GenEval and DPG-Bench, demonstrating that the reinforcement signal effectively enhances text–image semantic alignment. Excluding the confidence map slightly reduces overall accuracy, particularly in color and attribute precision, highlighting its role in fine-grained control.

Interestingly, removing the latent anchor leads to a larger performance drop. This is consistent with its purpose of design: the anchor provides structural and semantic priors that help stabilize latent representations and improve compositional understanding. Without it, the model becomes less robust on layout- and relation-sensitive tasks (e.g., multi-object or counting scenarios). These consistent trends across components confirm that each module contributes complementarily to the overall generation quality.

Training remained stable across all runs without oscillation in reward or divergence in PPO updates. 
This stability can be attributed to operating in the latent space, where the optimization landscape is smoother, and to the use of confidence-weighted rewards that dynamically adjust signal strength based on semantic reliability. 
Together, these mechanisms help maintain consistent convergence across different backbones and datasets.

\begin{table}[t]
\centering
\caption{Ablation study results across benchmarks.}
\label{tab:ablation_results}
\vspace{-.1in}
% 调整表格左右边距（列间距）
\setlength{\tabcolsep}{5pt} 

{\footnotesize
\begin{tabular}{lcc}
\toprule
Method Variant & GenEval (\%) & DPG-Bench (\%) \\
\midrule
RL(CLIPScore) & 78.78 & 84.15 \\
RL(Sentence-BERT) & 78.42 & 84.23 \\
Without RL & 77.68 & 83.84 \\
Without Confidence Map & 77.94 & 84.21 \\
Without Latent Anchor & 76.90 & 83.56 \\
\bottomrule
\end{tabular}
}
\vspace{-.2in}
\end{table}

\subsection{Discussion}
\label{sec:discussion}

\method is lightweight, operating entirely in latent space with the backbone frozen. On a single NVIDIA H100 GPU, training with a batch size of 8 takes about 1–2 seconds per batch (7–8 minutes per epoch), completing 15 epochs in roughly 2 hours. The latent reward projector adds minimal overhead, and PPO updates remain stable without additional gradient steps on the backbone. During inference, \method applies a single latent correction $\Delta_\theta$, requiring no reward computation or extra sampling, thus maintaining the same runtime as the base generator.

Compared to post-training approaches (e.g. HermesFlow~\citep{yang2025hermesflow}, UniRL~\citep{mao2025unirl}), which fine-tune large portions of the diffusion backbone, \method is far more parameter-efficient. Post-training typically updates hundreds of millions of parameters and incurs high compute costs across the full denoising trajectory. In contrast, our transformer predicts a latent correction tensor (e.g., $[1,16,128,128]$ for OmniGen2 \citep{wu2025omnigen2}) with fewer than 50M trainable parameters(generally under 1\% of the base model), yielding faster convergence, lower memory use, and greater stability.

In summary, while post-training methods reshape the generative distribution via large-scale fine-tuning, \method achieves comparable semantic and compositional gains through a localized latent correction that is both compute-efficient and backbone-independent. This suggests a promising direction for improving alignment through compact latent reasoning rather than full-model post-training.

\noindent\textbf{Influence and Broader Impact.}
Latent-level reinforcement correction offers an architecture-agnostic, plug-and-play enhancement for improving T2I models without retraining. It generalizes to diffusion, autoregressive, and even non-visual modalities (e.g., audio) where semantic consistency is essential. The interpretability analysis (Section~\ref{sec:interpretability}) also provides a means to understand and visualize language–latent interactions, aiding model diagnosis and human–AI co-creation. We expect latent alignment correction to inspire future research on intrinsically interpretable generative models.

\vspace{-.1in}
\section{Conclusion}
\label{sec:conclusion}

\vspace{-.1in}
In this work, we introduced \method, a general and interpretable framework for improving text-to-image generation through latent alignment and correction. By leveraging understanding-guided reinforcement signals in the latent space, our approach effectively bridges the gap between textual comprehension and visual generation. Extensive experiments across multiple benchmarks and backbones demonstrate consistent improvements in semantic fidelity, compositional reasoning, and interpretability.
Beyond quantitative gains, \method offers qualitative insights into how language concepts shape generative behavior, providing a step toward more controllable and explainable multimodal models. Future extensions may further enhance efficiency, reward design, and cross-domain adaptability, paving the way for transparent and human-aligned generative systems.

\noindent\textbf{Limitations and Future Work.}
Limitations include dependency on reward functions that may not capture aesthetic or cultural nuances, and interpretability signals that reflect trends rather than exact causality. Moreover, our study focuses on English prompts and common benchmarks. Future work will pursue more efficient correction strategies, human-aligned reward functions, and extensions to multilingual or dynamic generative tasks, further advancing controllable and explainable generation.
% \clearpage

\section*{Acknowledgment}
This research is supported by the Singapore Ministry of Education Academic Research Fund Tier 2 (Award No. MOE-T2EP20125-0016), and the Lee Kong Chian Fellowships. We also highlight the computing supported by Modal Academic Compute Grant.

{
    \small
    \bibliographystyle{ieeenat_fullname}
    \bibliography{main}
}

\setcounter{page}{1}
\maketitlesupplementary

\etocdepthtag.toc{mtappendix}
\etocsettagdepth{mtchapter}{none}
\etocsettagdepth{mtappendix}{subsection}

\tableofcontents

\appendix

\section{Method Details}

\subsection{Transformer-Based Corrector Architecture}

The Understanding-Guided Reinforcement Corrector (URC) is implemented as a lightweight transformer that operates on the latent feature grid of the frozen text-to-image model.  
Given a latent representation $z_0 \in \mathbb{R}^{C \times H \times W}$ and the prompt embedding $e_p$, URC predicts a residual $\Delta_\theta(z_0, e_p)$ applied before decoding.

\paragraph{Latent Tokenization.}
We flatten the spatial dimensions and treat each spatial location as a token: $Z_0 = \text{reshape}(z_0) \in \mathbb{R}^{(HW) \times C}$.  
2D sine–cosine positional encodings are added to retain spatial structure.

\paragraph{Prompt Conditioning.}
The prompt embedding $e_p \in \mathbb{R}^{d}$ is projected to the latent dimension via $e_p' = W_p e_p$ and integrated using Feature-wise Linear Modulation (FiLM):
\[
Z_0' = \gamma(e_p') \odot Z_0 + \beta(e_p'),
\]
where $\gamma$ and $\beta$ are produced from $e_p'$ through small MLPs. This allows the transformer to modulate latent tokens according to the semantic content of the prompt.

\paragraph{Transformer Layers.}
URC consists of six transformer blocks, each containing multi-head self-attention with 8 heads, cross-attention to prompt tokens, feedforward networks (hidden size $4C$), and pre-norm residual connections.  
Cross-attention explicitly injects linguistic structure, including object categories, colors, and relational phrases.

\paragraph{Residual Prediction.}
The output of the transformer is projected via a linear layer $W_o$ back to latent dimensionality, reshaped to form the residual:
\[
\Delta_\theta(z_0, e_p) \in \mathbb{R}^{C \times H \times W}.
\]
Despite its transformer architecture, URC remains compact ($\leq$ 15M parameters), ensuring it refines semantics without overpowering the frozen generator.

\subsection{Transformer-Based Latent Reward Projection}

The latent reward projector $R_\phi$ is a transformer that maps corrected latent activations and prompt embeddings to interpretable rewards approximating image-level feedback.

\paragraph{Input Construction.}
$R_\phi$ receives the corrected latent tokens $Z_c \in \mathbb{R}^{(HW) \times C}$, the prompt token embeddings $\{e_{p,i}\}$, and the CMD-derived global semantic vector $g_{\text{cmd}} \in \mathbb{R}^d$, which is appended as an extra token:
\[
X_0 = [Z_c; e_{p,1}; \dots; e_{p,T}; g_{\text{cmd}}].
\]

\paragraph{Transformer Design.}
$R_\phi$ has four transformer layers with 8-head multi-head attention, alternating self-attention and cross-attention, feedforward networks of size $4C$, and rotary positional embeddings for prompt tokens.

\paragraph{Reward Heads.}
After the transformer, the updated semantic token $g_{\text{cmd}}'$ is passed through three linear layers to produce the latent reward vector:
\[
r_\text{latent} = \big[ W_{\text{count}} g_{\text{cmd}}', \; W_{\text{color}} g_{\text{cmd}}', \; W_{\text{pos}} g_{\text{cmd}}' \big] \in \mathbb{R}^3,
\]
corresponding to counting, color, and position sub-rewards.

\paragraph{Training Objective.}
The projector is trained to minimize the L2 distance between predicted latent rewards and image-level task-specific rewards:
\[
\mathcal{L}_{\text{proj}} = \sum_{i=1}^{3} \| r_{\text{latent}}^{(i)} - r_{\text{image}}^{(i)} \|_2^2,
\]
enabling gradient-based optimization of URC even when the original image-level reward is non-differentiable.

\section{Additional Qualitative Results}

\subsection{Text-to-Image Generation}

\begin{figure*}[t!]
    % \vspace{-.25in}                                                
    \centering
    \includegraphics[width=.95\linewidth]{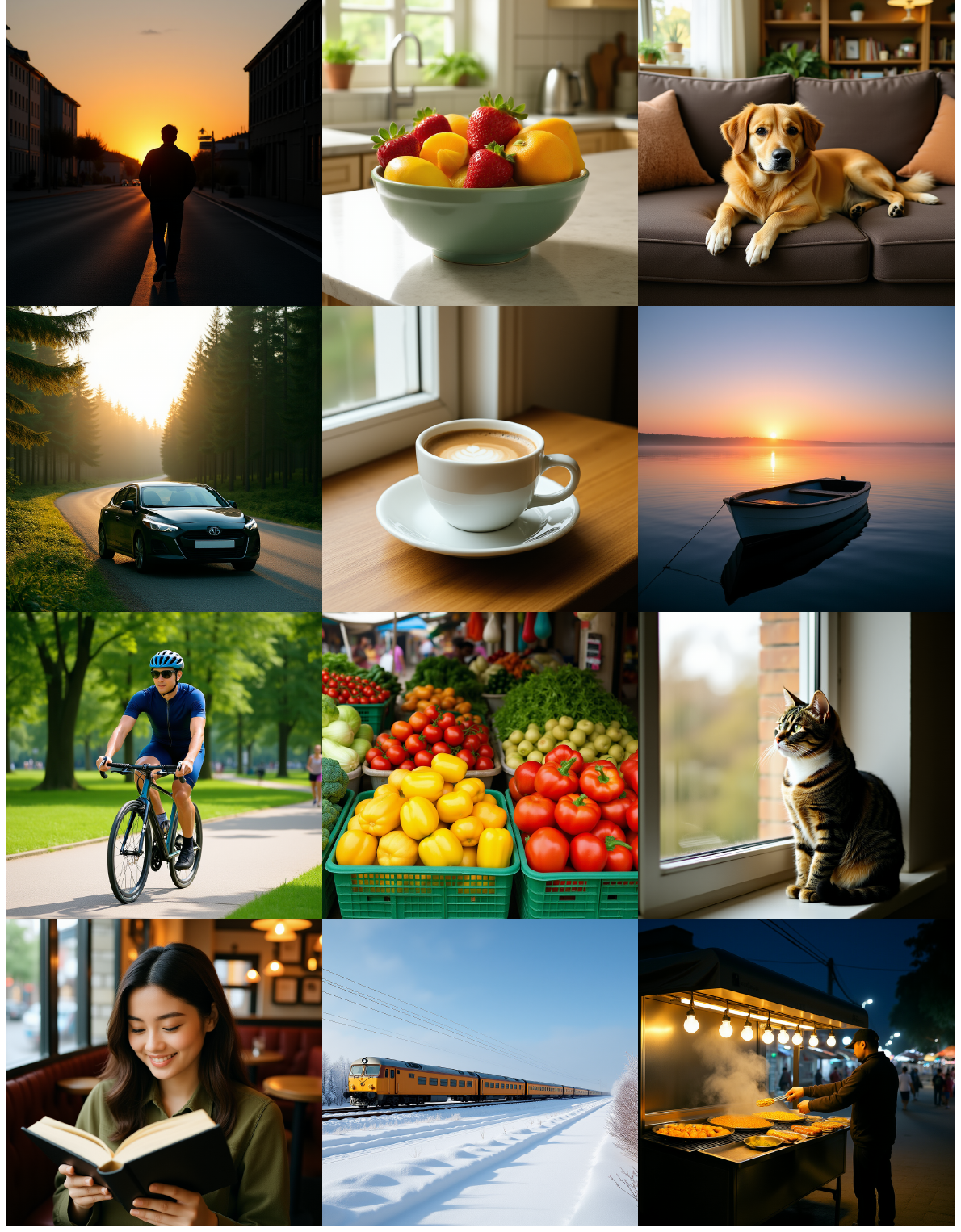}
    \vspace{-.6in}
    \caption{\textbf{Qualitative results for text to image generation.}}
    \label{fig:qualitative_results}
    \vspace{-.20in}
\end{figure*}

Additional qualitative results for text-to-image generation using our model are shown in Fig.\ref{fig:qualitative_results}. The prompts used for text-to-image generation, starting from top-left and going row-wise from left to right, are as follows:

\begin{enumerate}
    \item A person walking alone on a quiet street at sunset.
    \item A bowl of fresh fruit sitting on a kitchen counter.
    \item A dog lying on a couch in a cozy living room.
    \item A car parked beside a forest road in the morning.
    \item A cup of coffee on a wooden table near a window.
    \item A small boat floating on a calm lake at dawn.
    \item A cyclist riding through a city park.
    \item A marketplace stall filled with colorful vegetables.
    \item A cat sitting on a windowsill looking outside.
    \item A person reading a book in a quiet café.
    \item A train passing through a snowy landscape.
    \item A street food vendor cooking at night.
\end{enumerate}

\begin{figure*}[h]
    % \vspace{-.3in}
    \centering
    % Top image
    \includegraphics[width=\linewidth]{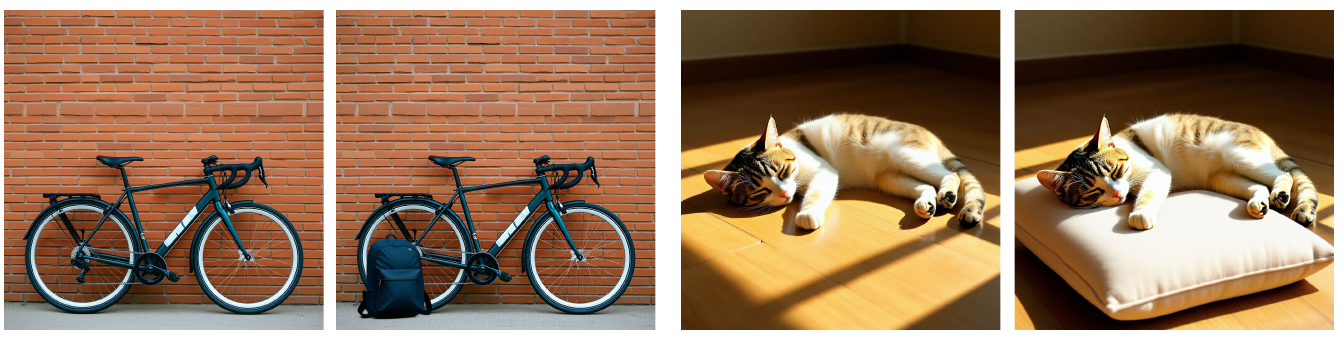}
    \vspace{0.5em}
    
    \vspace{-.28in}
    
    % Bottom image
    \includegraphics[width=\linewidth]{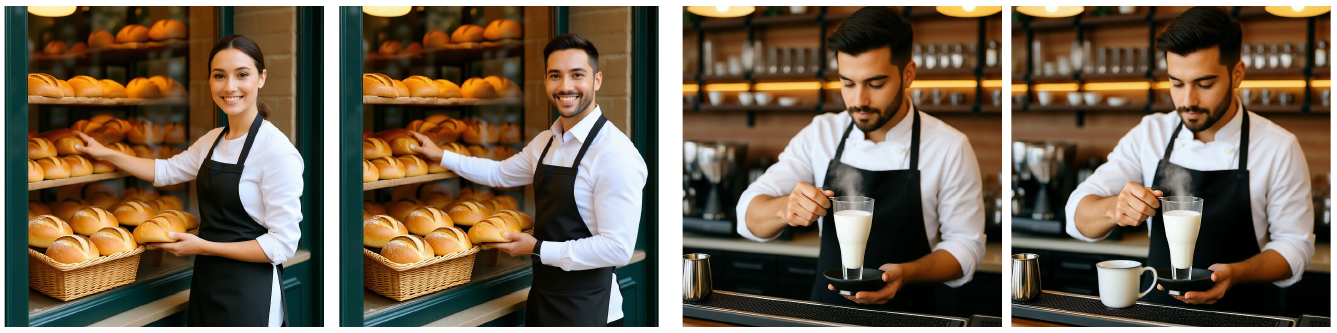}
    \vspace{0.5em}

    \vspace{-.28in}
    
    % Bottom image
    \includegraphics[width=\linewidth]{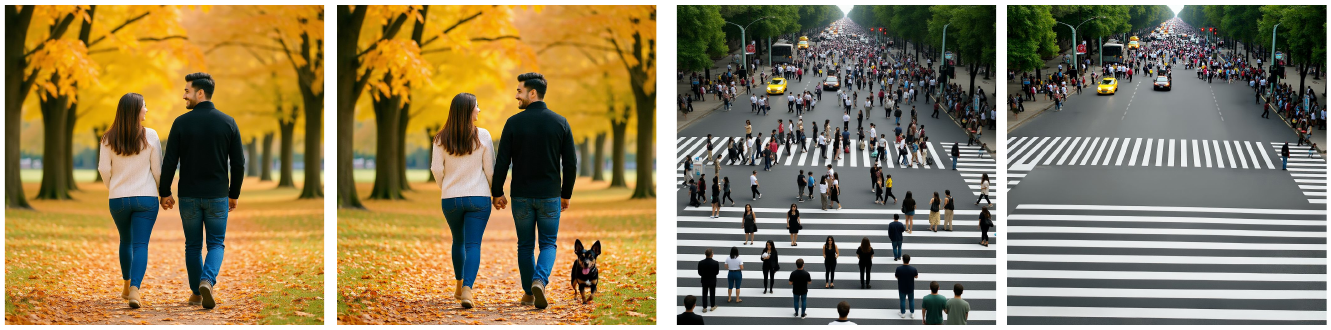}
    \vspace{0.5em}

    \vspace{-.28in}
    
    % Bottom image
    \includegraphics[width=\linewidth]{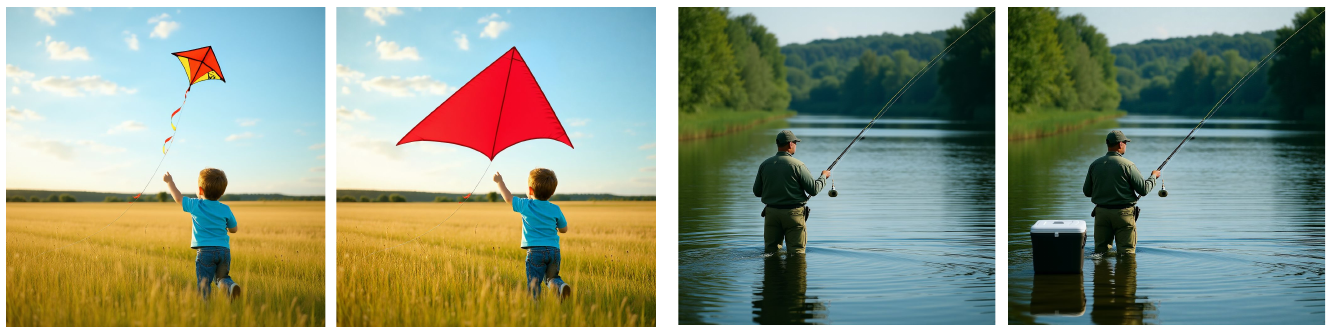}
    \vspace{0.5em}

    \vspace{-.28in}
    
    % Bottom image
    \includegraphics[width=\linewidth]{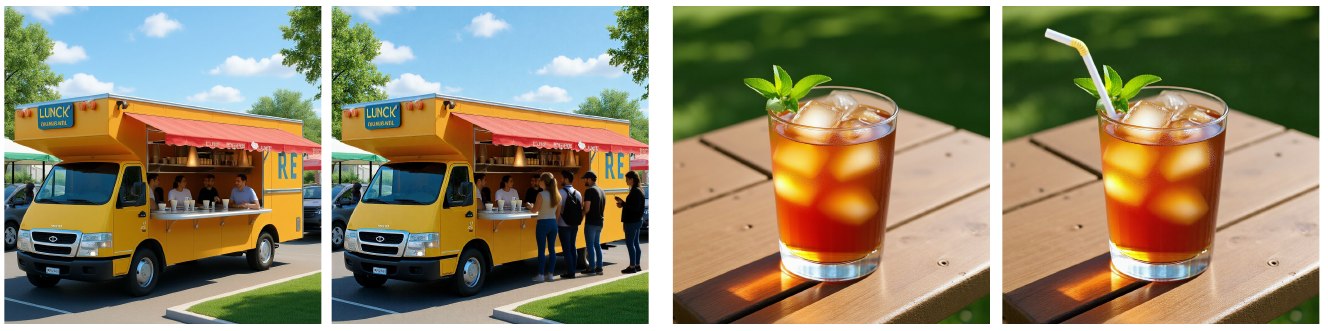}
    \vspace{0.5em}
    \vspace{-.28in}

    \vspace{-.1in}
    \caption{Qualitative editing results-1.}
    \label{fig:edit_examples1}
    \vspace{-.15in}
\end{figure*}

\begin{figure*}[h]
    % \vspace{-.3in}
    \centering
    % Top image
    \includegraphics[width=\linewidth]{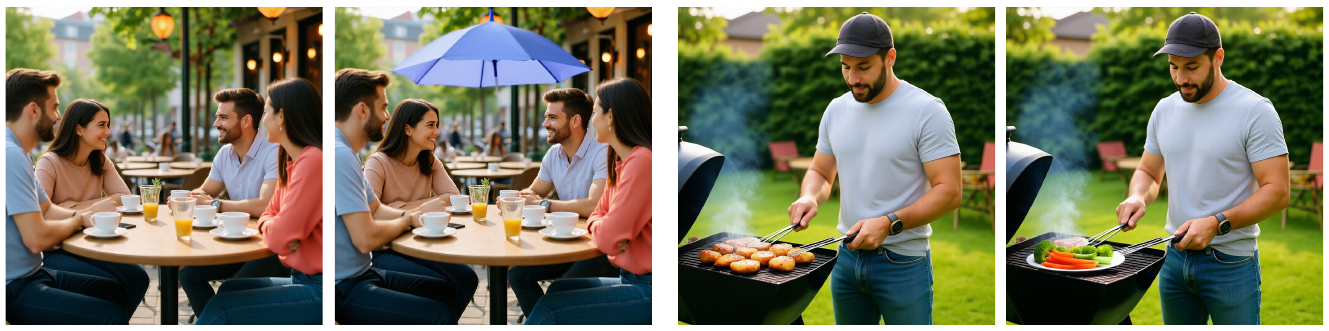}
    \vspace{0.5em}
    
    \vspace{-.28in}
    
    % Bottom image
    \includegraphics[width=\linewidth]{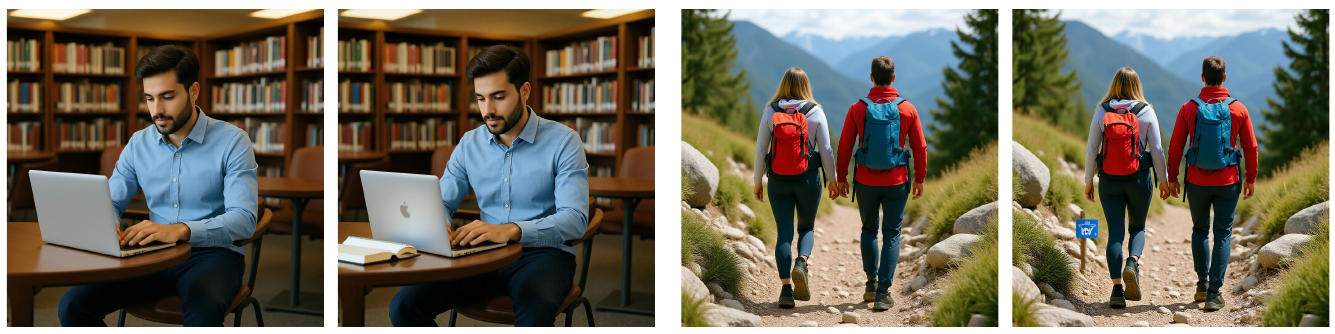}
    \vspace{0.5em}

    \vspace{-.28in}
    
    % Bottom image
    \includegraphics[width=\linewidth]{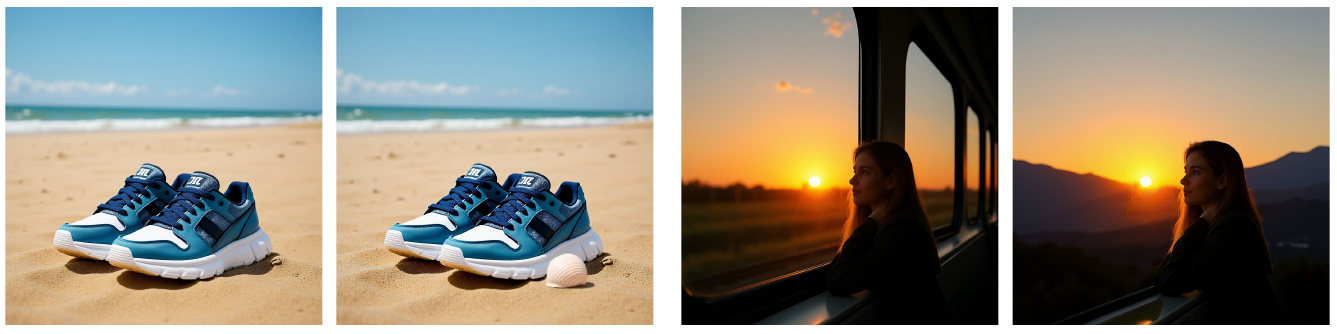}
    \vspace{0.5em}

    \vspace{-.28in}
    
    % Bottom image
    \includegraphics[width=\linewidth]{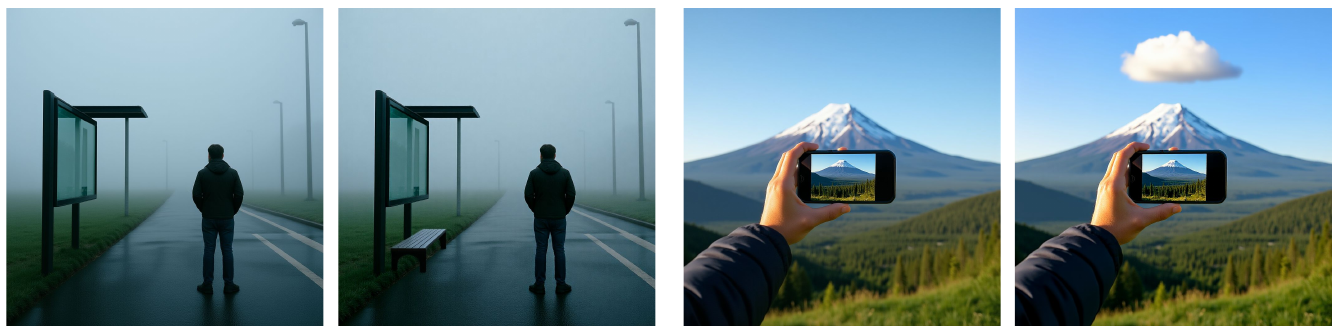}
    \vspace{0.5em}

    \vspace{-.28in}
    
    % Bottom image
    \includegraphics[width=\linewidth]{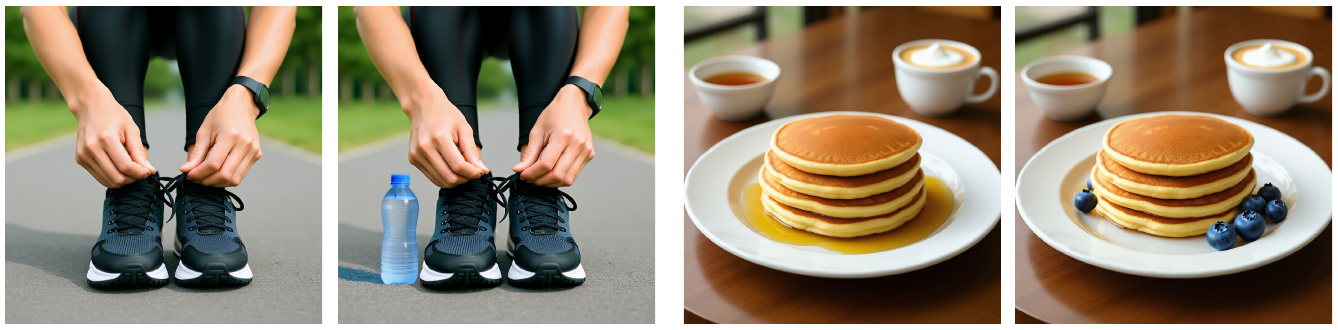}
    \vspace{0.5em}
    \vspace{-.28in}

    \vspace{-.1in}
    \caption{Qualitative editing results-2.}
    \label{fig:edit_examples2}
    \vspace{-.15in}
\end{figure*}

\begin{figure*}[h]
    \centering
    % Top image
    \includegraphics[width=\linewidth]{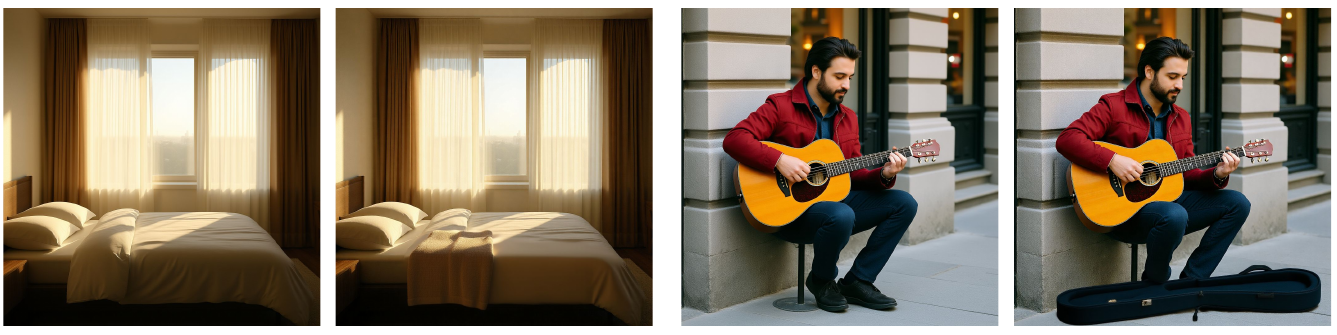}
    \vspace{0.5em}
    
    \vspace{-.28in}
    
    % Bottom image
    \includegraphics[width=\linewidth]{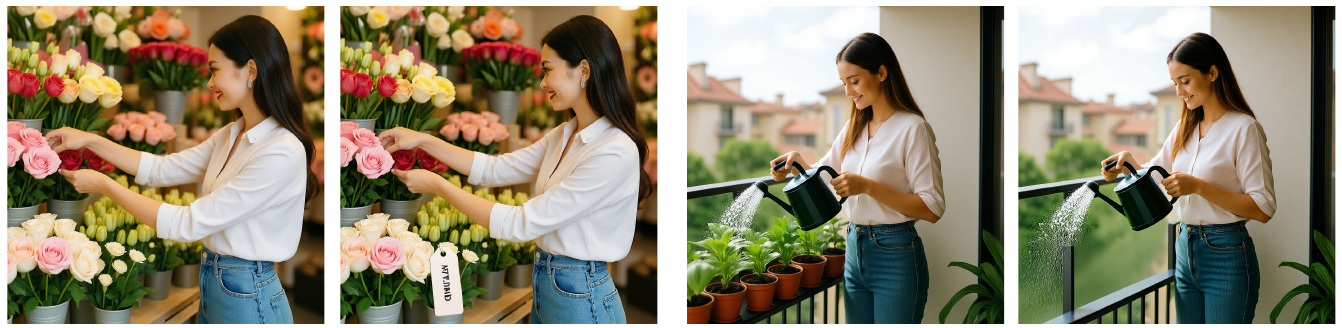}
    \vspace{0.5em}

    \vspace{-.1in}
    \caption{Qualitative editing results-3.}
    \label{fig:edit_examples3}
    \vspace{-.15in}
\end{figure*}

\subsection{Editing Tasks}

The prompts used to get results for the image editing task \Cref{fig:edit_examples1} \Cref{fig:edit_examples2} \Cref{fig:edit_examples3}, starting from the top-left and going row-wise from left to right, are as follows:
\begin{enumerate}
    \item Add a small backpack resting on the ground next to the bicycle.
    \item Add a small cushion under the cat.
    \item Turn the baker into a man.
    \item Add a small cup next to the pitcher.
    \item Add a small dog walking next to the couple.
    \item Remove half the people from the image, from the crosswalk.
    \item Change the kite’s color to bright red.
    \item Place a small cooler next to the fisherman, near his feet.
    \item Add a line of people near the food truck.
    \item Add a straw to the glass.
    \item Add an umbrella above the table.
    \item Add a plate of vegetables on the grill.
    \item Add a notebook next to the laptop.
    \item Add a small trail sign beside the path.
    \item Add a small seashell next to the sneakers.
    \item Add a distant mountain in the background scenery.
    \item Add a small bench next to the bus stop.
    \item Add a single cloud in the sky above the mountain.
    \item Add a small water bottle on the ground next to the person.
    \item Add a few blueberries on the plate beside the pancakes.
    \item Add a folded blanket at the side of the bed.
    \item Add an open guitar case on the ground in front of him.
    \item Add a price tag to one of the flower pots.
    \item Remove the plant pots from the row.
\end{enumerate}

\section{Supervised Data Generation}

Training prompts were generated using large language models to cover three semantic domains: color, position, and object count.  
For each domain, 10k candidate prompts were generated using multiple models (BLIP3o, SD3, and DEV). To ensure high-quality and semantically accurate prompts, all outputs were manually reviewed and filtered, retaining only those that correctly captured the intended attributes. Among the models tested, BLIP3o, SD3, and DEV consistently produced the most reliable and coherent prompts, showing strong consistency in describing colors, spatial relations, and object counts, whereas other models occasionally produced ambiguous or incomplete descriptions.  

Rather than using a separate testing set, the supervised dataset is used to guide the generator itself: for each validated prompt, the model generates multiple images using different random seeds. This self-generation process, supervised by the validated dataset, provides diverse latent representations and ensures coverage of the semantic domains, without requiring an explicit test split. The resulting dataset thus serves both as a source of training supervision and a reference for controlled evaluation during model development.

\section{Implementation and Reproducibility}

\subsection{Training Configuration}

We train using AdamW with a learning rate of $1{\times}10^{-4}$, batch size 8 per GPU, PPO clipping ratio 0.2, gradient clipping 1.0, and a cosine learning rate schedule. Experiments are conducted on H100 80GB GPUs.

\subsection{Hyperparameters}

The URC transformer has 6 layers, $R_\phi$ has 4 layers, embedding size 1024, residual scale $\alpha=0.8$, and task-weight modulation range $[0.5,2.0]$.

\section{Additional Experiments}

\noindent This section provides additional quantitative experiments complementing the analyses in the main paper.

\subsection{Process-Level Validation and Attribution}

\noindent \textbf{Causal Evidence.}
In Sec.~4.3.1, we demonstrate that improvements are directly driven by our formulation via three validation tests:
(i)~\textbf{Spatial Causal Link:} Masking high-activation LAM regions leads to a 6.3\% drop in CLIPScore, proving these regions are where alignment is fixed;
(ii)~\textbf{Signal-to-Gain Correlation:} A Spearman correlation of $\rho{=}0.71$ between token contribution magnitudes and reward gains confirms that latent rewards directly drive the correction;
(iii)~\textbf{Consistency:} A Jaccard similarity of 0.68 across prompts shows stable, predictable correction patterns.

\noindent \textbf{Incremental Dynamics.}
To quantify process-level dynamics, we analyzed semantic trajectories on a subset of 500 prompts from GenEval by sampling latent sub-rewards at irregular intervals through our differentiable projector $R_{\Phi}$. We observe a 26.34\% average reward increase between denoising steps 7 and 38, with a high Pearson correlation ($r{=}0.827$) between these incremental gains and final performance. Specifically, Counting rewards stabilize rapidly within the first 18 steps to establish structural layout, while Color and Position rewards provide continuous refinement through step 43. This temporal analysis confirms that xLARD performs active, multi-stage steering throughout the generative process rather than a singular post-hoc correction.

\subsection{Sub-Reward Analysis and Ablation}

Our modular design is a deliberate choice to ensure interpretability and avoid the black-box nature of aggregate rewards. While the sub-rewards are specific, they address the most frequent failure modes in T2I models and provide a framework that is easily extendable to other attributes. The ablation of reward roles is reported in Table~\ref{tab:ablation_expanded}.

\begin{table}[t]
\centering
\caption{Ablation study of reward components.}
\label{tab:ablation_expanded}
\vspace{-0.1in}
\setlength{\tabcolsep}{5pt}
{\footnotesize
\begin{tabular}{lcc}
\toprule
Method Variant & GenEval (\%) & DPG-Bench (\%) \\
\midrule
Full xLARD Framework & \textbf{81.24} & \textbf{86.72} \\ \midrule
w/o Counting Reward & 76.45 {\scriptsize($-$4.79)} & 83.10 {\scriptsize($-$3.62)} \\
w/o Color Reward    & 78.92 {\scriptsize($-$2.32)} & 84.55 {\scriptsize($-$2.17)} \\
w/o Position Reward & 79.15 {\scriptsize($-$2.09)} & 84.88 {\scriptsize($-$1.84)} \\
\bottomrule
\end{tabular}
}
\end{table}

\subsection{Comparison with Plug-and-Play Guidance Baseline}

For latent editing comparisons, we include a Plug-and-Play Guidance (PPGD) baseline adapted to our architecture; most existing latent editing methods are U-Net--based and do not generalize to unified multimodal transformers. On GenEval and DPG-Bench, PPGD achieves 77.04\% and 83.54\%, below our method (81.29\% and 86.45\%). These results provide additional comparative context against recent SOTA techniques.

\subsection{Confidence-Based Modulation (CMD) Analysis}

The Confidence Head $\omega$ is a lightweight MLP trained via PPO to predict sub-reward reliability. Rather than static weights, it acts as a dynamic gating mechanism to suppress irrelevant signals. Our analysis shows removing CMD causes a 3.3\% GenEval drop due to ``gradient interference,'' where unmodulated rewards compete for latent updates.

\subsection{Additional Technical Details}

\noindent \textbf{Reward Projection.}
The differentiable projector $R_\phi$ does not backpropagate through the decoder. Instead, it is trained via supervised regression to approximate non-differentiable image-level rewards, enabling stable gradient flow entirely within latent space during corrector optimization.

\noindent \textbf{Corrector Behavior.}
The corrector $\Delta_\theta$ is explicitly constrained to produce small-magnitude residual updates via the scaling factor $\alpha$ and PPO regularization, ensuring localized semantic refinement rather than latent overwriting or re-generation.

\noindent \textbf{Counting Implementation.}
Connected-component analysis is applied after adaptive thresholding and morphological filtering of token attention maps, which removes spurious activations and yields stable object-count estimates across prompts.

\subsection{Runtime and Memory}

Training takes 4\,h/epoch, inference matches base-generator runtime as mentioned in the paper discussion, and peak memory is approximately 72\,GB at batch size~8. Reference-prompt construction details are provided in the supplementary material.

% 
% To split the supplementary pages from the main paper, you can use \href{https://support.apple.com/en-ca/guide/preview/prvw11793/mac#:~:text=Delete%20a%20page%20from%20a,or%20choose%20Edit%20%3E%20Delete).}{Preview (on macOS)}, \href{https://www.adobe.com/acrobat/how-to/delete-pages-from-pdf.html#:~:text=Choose%20%E2%80%9CTools%E2%80%9D%20%3E%20%E2%80%9COrganize,or%20pages%20from%20the%20file.}{Adobe Acrobat} (on all OSs), as well as \href{https://superuser.com/questions/517986/is-it-possible-to-delete-some-pages-of-a-pdf-document}{command line tools}.

\end{document}